\documentclass{article}

\PassOptionsToPackage{numbers,{sort&compress}}{natbib}

\usepackage[main, final]{neurips_2026}
\usepackage{amsmath,amssymb,amsfonts}
\usepackage{subcaption}

\usepackage[utf8]{inputenc} 
\usepackage[T1]{fontenc}    
\usepackage{hyperref}       
\usepackage{url}            
\usepackage{booktabs}       
\usepackage{amsfonts}       
\usepackage{nicefrac}       
\usepackage{microtype}      
\usepackage{xcolor}         
\usepackage{multirow}
\usepackage{graphicx}
\usepackage{enumitem}

\title{Beyond Task-Agnostic: Task-Aware Grouping for Communication-Efficient Multi-Task MoE Inference}

\author{%
  Zhiyao Xu\textsuperscript{1}\quad
  Aoxue Liu\textsuperscript{1}\quad
  Zhanjie Ding\textsuperscript{1}\quad
  Dan Zhao\textsuperscript{2}\footnotemark[1]\quad
  Yong Jiang\textsuperscript{1, 2}\footnotemark[1]\thanks{Corresponding author.} \quad
  Qing Li\textsuperscript{2}
  \\[0.4em]
  \textsuperscript{1}Tsinghua Shenzhen International Graduate School, Shenzhen, China \qquad
  \\
  \textsuperscript{2}Pengcheng Laboratory, Shenzhen, China
  \\[0.2em]
  \texttt{\{xu-zy25, lax25, dingzj25\}@mails.tsinghua.edu.cn} \quad
  \\
  \texttt{\{zhaod01, liq\}@pcl.ac.cn} \quad
  \texttt{jiangy@sz.tsinghua.edu.cn}
}

\begin{document}

\maketitle

\begin{abstract}
  Sparsely activated Mixture-of-Experts (MoE) models scale capacity via conditional computation, but distributed inference suffers from cross-GPU expert communication and routing-induced load imbalance. Existing placement methods reduce this cost by co-locating frequently co-activated experts; however, they derive a single deployment plan from globally aggregated routing traces, thereby averaging away the heterogeneous, task-specific co-activation patterns that actually drive communication in multi-task serving. We observe that expert co-activation is strongly task-conditioned: pairs tightly coupled in one task family are often uncorrelated in another, so effective deployment should group experts by task-aware co-activation rather than by a task-agnostic average. Based on this insight, we propose \emph{Task-Aware Coactivation Grouping} (TACG), a deployment-time framework that uses family-specific dispatch and co-activation traces to derive per-expert task-family preferences, reweights the co-activation graph so that intra-family locality dominates grouping, and assigns each expert to a primary GPU under exact capacity constraints. To keep the static placement robust under online workload skew, we further introduce \emph{Generic Expert Shared Replication} (GESR), a lightweight companion that identifies generic experts with consistently central co-activation profiles, replicates them across a small set of secondary GPUs, and applies locality- and load-aware selection at serving time. Experiments on three representative open-source MoE models demonstrate that our framework reduces the average communication cost by 31.39\% over the baseline, while preserving an average Jain fairness index of 0.9975. This advantage persists even under severe distribution shifts in the inference data, consistently outperforming strong baselines. 
  
  
\end{abstract}

\section{Introduction}
\label{sec:intro}


By leveraging conditional computation, MoE can significantly increase model capacity without a proportional increase in floating-point operations, and has therefore been widely adopted in all kinds of models~\citep{shazeer2017outrageously, lepikhin2021gshard, fedus2022switch}. Despite this advantage, MoE models typically impose substantially higher memory demands than their dense counterparts, often exceeding the capacity of a single accelerator.  Consequently, large-scale MoE inference is commonly deployed using a combination of expert parallelism and data parallelism~\citep{lepikhin2021gshard, gale2023megablocks, hwang2023tutel}.  In this setting, experts in each MoE layer are distributed across multiple GPUs and coordinated through All-to-All communication.  Although this distributed design enables scalable deployment, it also introduces two key system challenges: excessive communication overhead and severe load imbalance. \citep{li2026semantic, bambhaniya2026scalingmultinodemixtureofexpertsinference} show that the all-to-all communication overhead generated by the expert distribution accounts for 30\%-50\% of the overall reasoning time overhead of MoE.

Existing MoE serving optimizations mitigate these costs from several complementary directions, including more efficient sparse kernels, improved All-to-All scheduling, expert placement, and dynamic replication~\citep{MLSYS2023_5616d34c, MLSYS2023_5a54f793, 10.1145/3600006.3613165, 288705, 10579139}. Among them, expert placement is particularly attractive because it can reduce communication without altering the model architecture: if experts that are frequently co-activated are placed on the same GPU, each token contacts fewer remote devices during inference. 

\begin{figure}[h]
    \centering
    \begin{subfigure}[b]{0.32\linewidth}
        \centering
        \includegraphics[width=\linewidth]{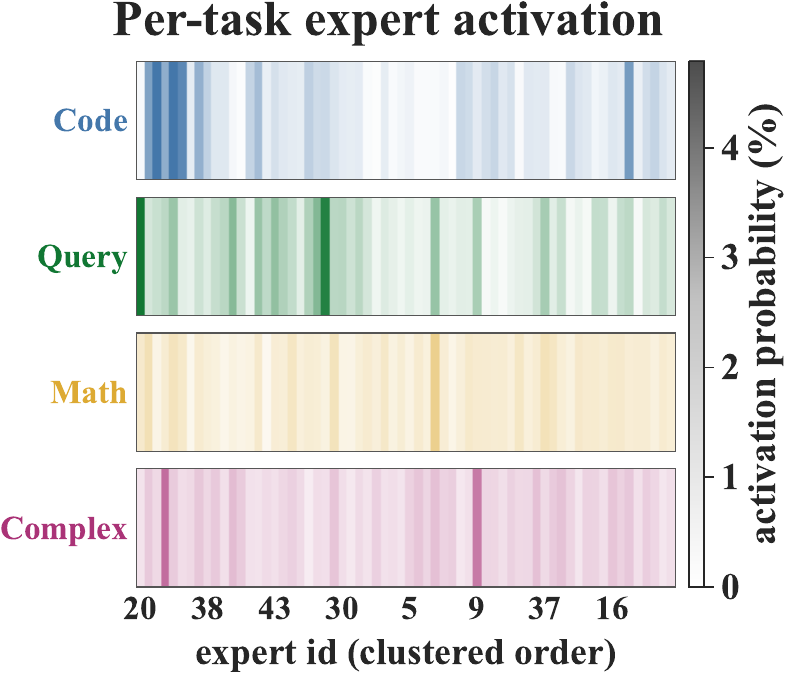}
        \label{fig:1}
    \end{subfigure}
    \hfill
    \begin{subfigure}[b]{0.32\linewidth}
        \centering
        \includegraphics[width=\linewidth]{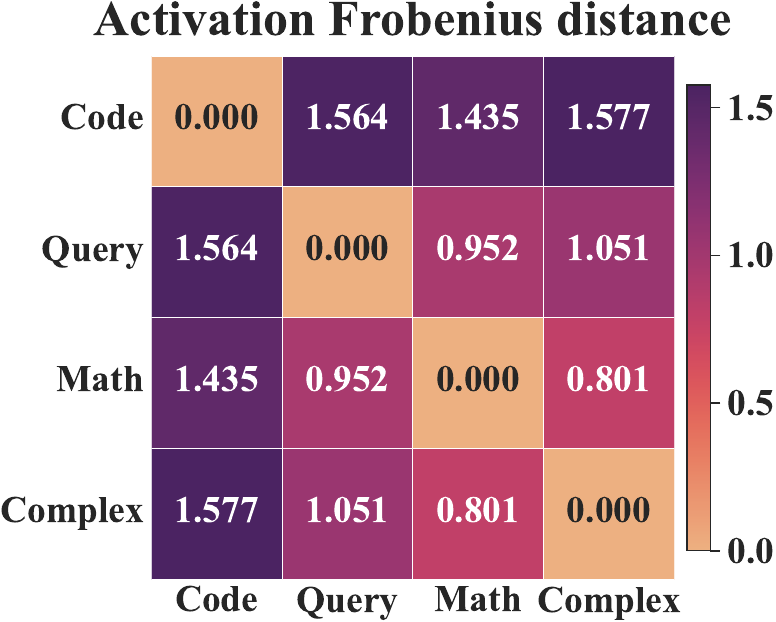}
        \label{fig:2}
    \end{subfigure}
    \hfill
    \begin{subfigure}[b]{0.32\linewidth}
        \centering
        \includegraphics[width=\linewidth]{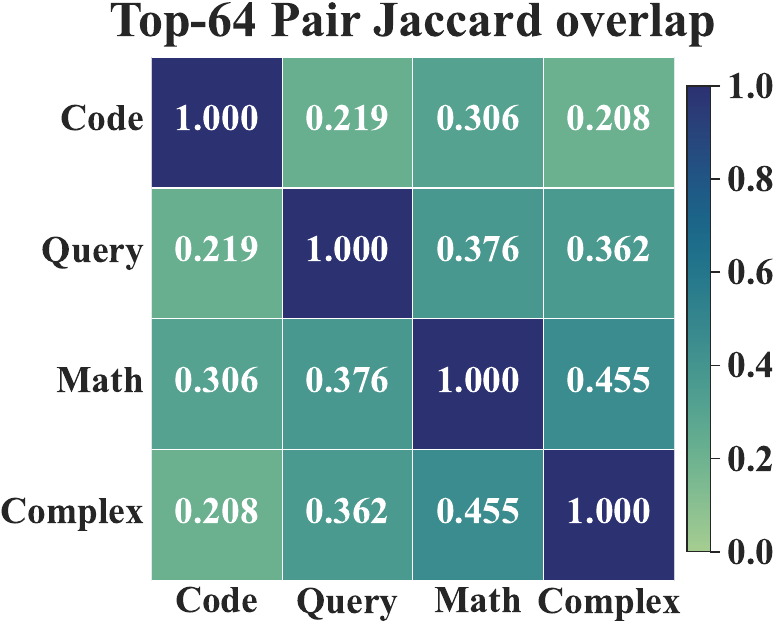}
        \label{fig:3}
    \end{subfigure}
    \caption{Different tasks yield significantly distinct expert activations (left), quantified by Frobenius distance (middle), while their induced expert collaboration patterns rarely overlap (right).}
    \label{fig:three re}
\end{figure}

However, most existing strategies~\citep{zhang-etal-2025-advancing, go2025moetuneroptimizedmixtureexpert, han2026gracemoegroupingreplicationlocalityaware} estimate expert affinity from globally aggregated routing traces and then construct a single deployment plan shared by all incoming requests. Such a task-agnostic view is natural for homogeneous workloads, but it is increasingly misaligned with modern shared inference services that interleave multiple task (code, math, query, complex reasoning, etc.) requests. Averaging heterogeneous coactivation patterns into one graph collapses task-specific expert groups and leaves a substantial amount of communication locality unexploited, making this kind of grouping difficult to be universal.


Our key insight is that expert coactivation is strongly task-conditioned. As shown in Figure~\ref{fig:three re}, two experts that are tightly coupled on one task family are often only weakly related on another, or even competing for capacity. This observation motivates us to shift the unit of coactivation analysis from the workload to task awareness. At the same time, once the actual reasoning data is shifted in distribution after deployment, the prior grouping without task awareness will not work properly because different task types often activate different expert collaboration patterns.

So we propose \textbf{Task-Aware Coactivation Grouping (TACG)}, a deployment-time framework that converts task-conditioned coactivation into a capacity-feasible placement. It derives soft expert–family preferences from traces, constructs a task-modulated coactivation graph, making task structure an explicit driver of locality. To protect these locality gains against load skew caused by a few generic experts that become hotspots under workload shift, we pair TACG with a lightweight \textbf{Generic Expert Shared Replication (GESR)} module, which identifies such experts via coactivation centrality and cross-family consistency, assigns each a small set of secondary GPUs chosen from structurally compatible non-primary cards, and selects replicas at serving time using a load-aware rule with exponentially decayed load signals.

We evaluate the proposed framework on three representative MoE models. Compared with the baseline placement, our method reduces the average communication metric by 31.39\% across models, consistently improving over strong placement strategies. Meanwhile, this reduction is obtained while maintaining competitive load balance: our average Jain index remains 0.9975 and the average maximum load violation is 0.0736. Ablation studies confirm that the task-aware coactivation signal drives most of the gain, with GESR keeping load balance under control.



Our contributions are summarized as follows:
\begin{itemize}[leftmargin=*]
    \item We identify that expert coactivation in multi-task MoE serving is strongly task-conditioned, and argue that an effective expert placement should be driven by task-aware coactivation rather than a single globally averaged affinity graph.
    \item We propose \textbf{Task-Aware Coactivation Grouping (TACG)}, a task-aware strategy that reweights the coactivation graph by expert-to-family preferences and produces a capacity-feasible placement that strictly generalizes task-agnostic grouping.
    \item We propose \textbf{Generic Expert Shared Replication (GESR)}, a lightweight companion that selectively replicates cross-family generic experts and performs load-aware online guard.
    \item We demonstrate on three representative MoE models that the framework delivers consistent communication reduction while retaining strong load-balance metrics, with ablations isolating the contribution of task-aware coactivation as the dominant factor.
\end{itemize}

\section{Related Work}
\label{sec:related_work}

\subsection{MoE Distributed Architectures and Expert Placement}

Sparse Mixture-of-Experts architectures increase model capacity by routing each token to a small subset of experts. GShard~\citep{lepikhin2021gshard} and Switch Transformer~\citep{JMLR:v23:21-0998} established the dominant sparse-routing paradigm for large-scale Transformers, while BASE Layers~\citep{pmlr-v139-lewis21a} and Expert Choice routing~\citep{zhou2022mixtureofexperts} improve utilization and training stability through globally balanced assignment and expert-driven token selection. DeepSeekMoE~\citep{dai-etal-2024-deepseekmoe} strengthens expert specialization via fine-grained expert partitioning and shared experts. A more closely related line studies expert placement itself. ExFlow~\citep{10579139} exploits inter-layer expert affinity for placement; C2R~\citep{zhang-etal-2025-advancing} uses collaboration-constrained routing to encourage specialized expert groups and reduce redundant communication; MoETuner~\citep{go2025moetuneroptimizedmixtureexpert} formulates placement as an ILP that jointly considers communication, computation, and load balance; and Semantic Parallelism~\citep{li2026semantic} further reduces cross-group communication by semantically aligning token inputs to each group. Despite these advances, existing methods usually derive a single global placement from aggregated affinity, collaboration, or routing-dependency statistics, and do not explicitly analyze experts’ task preferences.



\subsection{Task-Aware MoE}

Task-aware MoE methods show that expert utility is often strongly conditioned on the task being served. MMoE~\citep{10.1145/3219819.3220007} and DSelect-k~\citep{10.5555/3540261.3542507} introduce task-specific gating for multi-task learning, demonstrating that different tasks benefit from different expert mixtures. Beyond Distillation~\citep{kudugunta-etal-2021-beyond-distillation} further shows that task-level routing can extract smaller and more efficient deployable sub-networks than token-level routing alone. Related pruning and adaptation methods, including Task-Specific Expert Pruning~\citep{chen2022taskspecificexpertpruningsparse}, Task-Based MoE~\citep{pham-etal-2023-task}, and HMoRA~\citep{ICLR2025_e43a3399}, also reveal that task information can be injected into routing or expert selection at different granularities. \citep{NEURIPS2025_4598de7d} improves the performance of MoE models by strengthening the experts’ task preferences through post-training with expert specialization. Recent research~\citep{bambhaniya2026scalingmultinodemixtureofexpertsinference} from Meta also points to the potential of task-specific grouping. However, these approaches mainly act during training, fine-tuning, adaptation, or pruning.

\section{Method}

\subsection{Overview}

\begin{figure}[h]
\centering
\includegraphics[width = \textwidth]{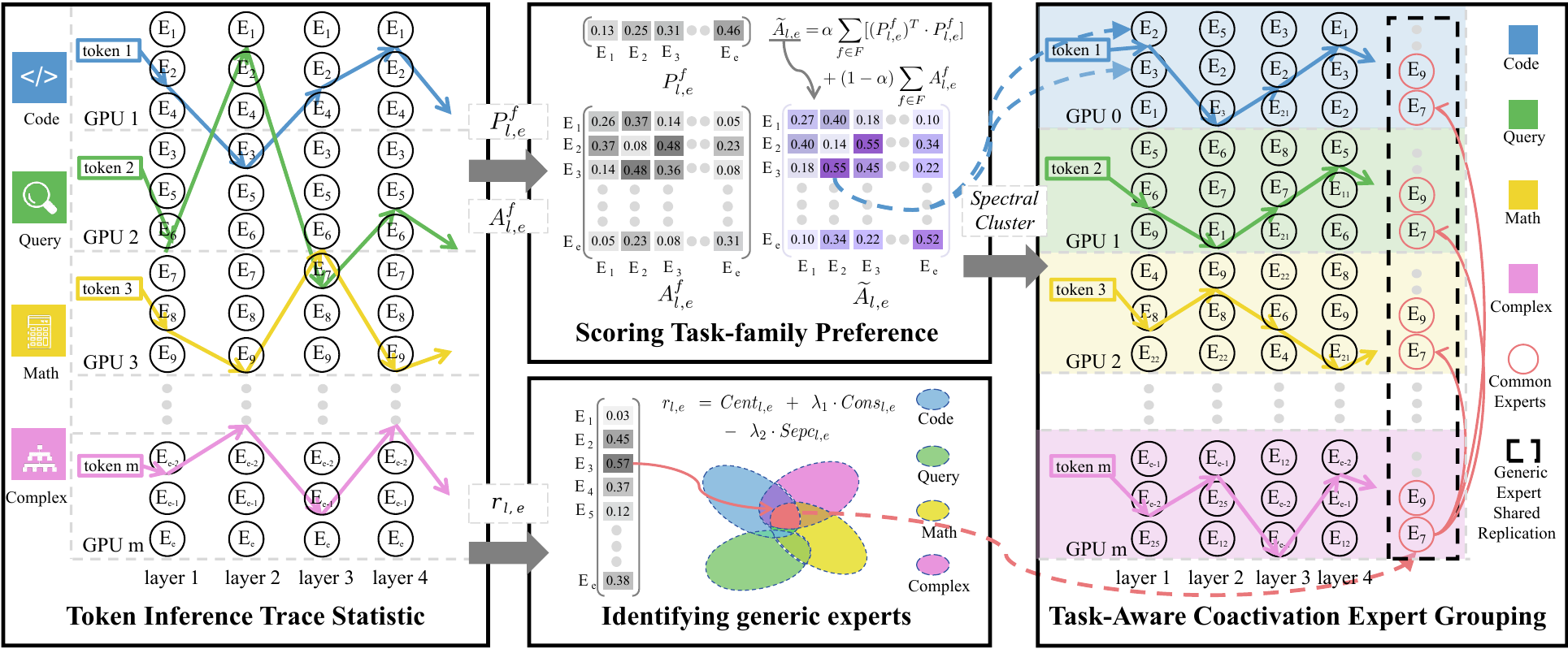}
\caption{Overview: From Default Grouping to Task-Aware Grouping}
\label{fig:Overview}
\end{figure}

As illustrated in Figure~\ref{fig:Overview}, our framework runs an offline pipeline followed by an online guard. \emph{Offline}, we first split calibration traffic by task family and collect per-family usage and coactivation statistics at every layer; we then score each expert's task-family preference and use it to reweight the coactivation graph into a task-modulated graph; finally, spectral clustering with a capacity-repair pass groups experts on this graph, producing one primary GPU per expert---these three steps form \textbf{Task-Aware Coactivation Grouping (TACG)}. \emph{Online}, \textbf{Generic Expert Shared Replication (GESR)} flags a small set of cross-family generic experts, attaches a few secondary GPUs to each, and at serving time picks one candidate per token via a locality- and load-aware rule, leaving the MoE router untouched.


\subsection{Problem Setup}


Consider an MoE model with $L$ MoE layers, $E$ experts per layer, and $M$ GPUs, where each GPU denotes the expert-parallel serving unit. The task set is partitioned into $K=4$ families. 
\begin{equation}
    \mathcal{F}=\{\texttt{code},\texttt{query},\texttt{math},\texttt{complex\_reasoning}\}.
    \label{eq:family-set}
\end{equation}

For each family $f\in\mathcal{F}$, we reserve a disjoint GPU set $\mathcal{G}_f$ with $\cup_f \mathcal{G}_f=\{1,\ldots,M\}$. Let $B_f$ be the number of experts nominally allocated to family $f$, with $\sum_f B_f=E$, and let $C_{f,m}$ be the capacity of GPU $m\in\mathcal{G}_f$, with $\sum_{m\in\mathcal{G}_f}C_{f,m}=B_f$. Given calibration routing traces grouped by task family, TACG outputs a primary placement $g_{l,e}^{\mathrm{primary}}$ for every expert; GESR then attaches a sparse candidate set $\mathcal{C}_{l,e}$ to a small subset of generic experts. The MoE router itself is unchanged; all decisions are made at deployment and serving time. Section~\ref{sec:primary_placement} details TACG, and Section~\ref{sec:shared_adaptation} details GESR.

\subsection{Task-Aware Coactivation Grouping (TACG)}
\label{sec:primary_placement}


\paragraph{Scoring task-family preference.}
We first quantify how strongly each expert serves each task family. For each layer $l$ and family $f$, we compute two statistics from calibration traces: the usage vector $u_l^{(f)}\in\mathbb{R}^{E}$ (relative dispatch frequency of each expert) and the coactivation matrix $A_l^{(f)}\in\mathbb{R}^{E\times E}$ (frequency of expert pairs selected by the same token). To avoid simply highlighting globally popular experts, we score each expert by its relative advantage for one family against the other families:
\begin{equation}
    \Delta u_{l,e}^{(f)}
    =
    u_{l,e}^{(f)}
    -
    \frac{1}{K-1}\sum_{f'\neq f}u_{l,e}^{(f')},
    \qquad
    c_{l,e}^{(f)}
    =
    \sum_{e'} \left|A_l^{(f)}(e,e')\right|,
    \label{eq:usage-coact-advantage}
\end{equation}
\begin{equation}
    \Delta c_{l,e}^{(f)}
    =
    c_{l,e}^{(f)}
    -
    \frac{1}{K-1}\sum_{f'\neq f}c_{l,e}^{(f')}.
    \label{eq:coact-advantage}
\end{equation}

We apply layer-wise z-score normalization across experts so that usage and coactivation advantages have comparable scales:
\begin{equation}
\operatorname{zscore}(x_{l,e}^{(f)})
=
\frac{x_{l,e}^{(f)}-\mu_l^{(f)}}{\sigma_l^{(f)}+\epsilon},
\label{eq:zscore}
\end{equation}
where $\mu_l^{(f)}$ and $\sigma_l^{(f)}$ are the mean and standard deviation of $x_{l,e}^{(f)}$ over experts $e$ in layer $l$ for family $f$, and $\epsilon$ is a small constant for numerical stability.

After applying this normalization to both $\Delta u$ and $\Delta c$, the soft preference score is
\begin{equation}
    s_{l,e}^{(f)}
    =
     \operatorname{zscore}(\Delta u_{l,e}^{(f)})
    +
     \operatorname{zscore}(\Delta c_{l,e}^{(f)}),
    \label{eq:soft-score}
\end{equation}
and a temperature-softmax converts this score into a soft task-family preference distribution:
\begin{equation}
    p_{l,e}^{(f)}
    =
    \frac{\exp(s_{l,e}^{(f)}/\tau)}
    {\sum_{f'\in\mathcal{F}}\exp(s_{l,e}^{(f')}/\tau)} .
    \label{eq:family-pref}
\end{equation}
The distribution $p_{l,e}^{(f)}$ preserves uncertainty for experts that are useful to multiple families. Crucially, TACG never converts $p_{l,e}^{(f)}$ into a rigid expert-to-family partition; it uses $p_{l,e}^{(f)}$ as a soft task signal that modulates the coactivation graph in the next step.

\paragraph{Task-modulated coactivation graph.}
Pooled coactivation $\bar{A}_l=\tfrac{1}{K}\sum_{f} A_l^{(f)}$ summarizes which experts are jointly accessed by a token, but it is task-agnostic: it treats coactivation edges across different families as equally valuable for locality. In multi-task serving, two experts that co-activate mainly on different families rarely meet on the same token, so placing them together wastes co-location capacity that would be better spent on same-family pairs. TACG therefore modulates the coactivation signal with the expert-to-family preference $p_{l,e}^{(f)}$ before grouping, so that the grouping step sees an affinity that favors expert pairs which both co-activate and tend to serve the same family.

Concretely, we stack the preferences into $P_l\in\mathbb{R}^{E\times K}$ with $(P_l)_{e,f}=p_{l,e}^{(f)}$, and form the \emph{same-family kernel}
\begin{equation}
    A_l^{\mathrm{task}} = P_l P_l^{\top} \in [0,1]^{E\times E},
    \qquad
    A_l^{\mathrm{task}}(e,e')=\sum_{f\in\mathcal{F}} p_{l,e}^{(f)}\, p_{l,e'}^{(f)},
    \label{eq:same-family-kernel}
\end{equation}
which is the probability that two experts are drawn from the same family under independent family affiliations. With $\hat{A}_l=\bar{A}_l/\max \bar{A}_l$ normalized to $[0,1]$, the task-modulated coactivation graph is
\begin{equation}
    \tilde{A}_l
    =
    (1-\alpha)\,\hat{A}_l
    +
    \alpha\,\bigl(A_l^{\mathrm{task}}\odot \hat{A}_l\bigr),
    \label{eq:task-modulated-graph}
\end{equation}
where $\odot$ denotes the Hadamard product and $\alpha\in[0,1]$ controls how strongly task structure reshapes coactivation. The multiplicative gate $A_l^{\mathrm{task}}\odot \hat{A}_l$ is the key design choice. It amplifies coactivation edges whose endpoints share a family and attenuates those that do not, without inventing new edges between experts that never co-activate. Thus $p_{l,e}^{(f)}$ enters grouping as a principled edge reweighting rather than a hard label.

\paragraph{Capacity-feasible expert grouping.}
Given $\tilde{A}_l$, we partition experts into $M$ exact-size groups $\{\mathcal{S}_{l,1},\ldots,\mathcal{S}_{l,M}\}$ by maximizing within-group affinity under per-GPU capacities:
\begin{equation}
    \max_{\{\mathcal{S}_{l,m}\}}
    \sum_{m=1}^{M}
    \sum_{e,e'\in\mathcal{S}_{l,m}}
    \tilde{A}_l(e,e')
    \quad
    \text{s.t.}\quad
    |\mathcal{S}_{l,m}|=C_m,\ \
    \bigsqcup_{m=1}^{M}\mathcal{S}_{l,m}=\{1,\ldots,E\}.
    \label{eq:grouping-objective}
\end{equation}
This capacity-constrained graph partitioning problem is NP-hard, so we solve it with a two-stage spectral-then-repair procedure that is cheap to run offline and, empirically, very close to exhaustive search on the reweighted graph.

\emph{Stage 1: spectral clustering.} We symmetrize $\tilde{A}_l$, scale it to $[0,1]$, and add a small diagonal jitter for numerical stability. Treating the result as a precomputed affinity, we apply normalized spectral clustering: we form the symmetric normalized Laplacian $L_{\mathrm{sym}}=I-D^{-1/2}\tilde{A}_l D^{-1/2}$ with degree matrix $D$, embed experts via its $M$ smallest eigenvectors, and run $k$-means with $k=M$ to obtain a soft partition $\{\mathcal{S}_{l,m}^{0}\}$. This step directly targets the relaxed within-group affinity objective and exposes the communities implied by the task-modulated graph, but it ignores the exact-size constraints $|\mathcal{S}_{l,m}|=C_m$.

\emph{Stage 2: capacity repair.} We enforce the per-GPU capacities with a local, affinity-guided repair that never re-runs spectral embedding. For every overfull group $m$ (with $|\mathcal{S}_{l,m}^{0}|>C_m$), we rank its members by their intra-group affinity $\sum_{e'\in\mathcal{S}_{l,m}^{0}}\tilde{A}_l(e,e')$ and keep the top-$C_m$; the remaining experts enter an overflow pool. Each overflow expert $e$ is then greedily inserted into the non-full group $m^{\star}$ that yields the largest increase in within-group affinity, i.e.
\begin{equation}
    m^{\star}(e)=\arg\max_{m:\,|\mathcal{S}_{l,m}|<C_m}\,\sum_{e'\in\mathcal{S}_{l,m}}\tilde{A}_l(e,e').
    \label{eq:greedy-insert}
\end{equation}
After this first pass, any residual size mismatches are resolved by a donor--receiver swap loop: for each still-underfull group $\mathcal{S}_{l,\mathrm{dst}}$, we take the weakest member (lowest intra-group affinity) of an overfull donor group and move it to $\mathcal{S}_{l,\mathrm{dst}}$ only if the move improves the sum of within-group affinities. The loop terminates when all groups satisfy $|\mathcal{S}_{l,m}|=C_m$, which is guaranteed within a bounded number of iterations since every accepted move strictly raises a monotone bounded objective.

In our experiments we fix $\alpha=0.25$, which keeps communication as the dominant signal while gating out cross-family edges that hurt multi-task locality. The resulting assignment defines the primary placement $g_{l,e}^{\mathrm{primary}}$, which co-locates stable task-coherent expert communities and strictly generalizes task-agnostic grouping.


\subsection{Generic Expert Shared Replication (GESR)}
\label{sec:gesr}
\label{sec:shared_adaptation}

TACG delivers a stable and communication-efficient primary placement, but a small fraction of experts are \emph{generic}: their coactivation profile is consistently central across families, so they are dispatched under every task and can become load hotspots. Replicating many experts would inflate memory cost and possibly create new hotspots, while a dedicated common-GPU would reintroduce exactly the cross-family traffic TACG eliminates. GESR addresses this risk with a minimal intervention on top of TACG: only a small set of generic experts receives secondary GPUs, and online selection is restricted to this candidate set, using TACG's own coactivation structure as the compatibility criterion.

\paragraph{Identifying generic experts.}
We identify generic experts using a score that favors global centrality and cross-family consistency while penalizing strong task-specific deviation. For expert $e$ at layer $l$ with pooled coactivation $\bar{A}_l$,
\begin{equation}
    r_{l,e}
    =
    \underbrace{\mathrm{Cent}_{l,e}}_{\text{centrality}}
    +
    \lambda_1
    \underbrace{\mathrm{Cons}_{l,e}}_{\text{cross-family consistency}}
    -
    \lambda_2
    \underbrace{\mathrm{Spec}_{l,e}}_{\text{task-specific penalty}},
    \label{eq:generic-score}
\end{equation}
where $\mathrm{Cent}_{l,e}=\sum_{e'}\bar{A}_l(e,e')$, $\mathrm{Cons}_{l,e}$ is the average cosine similarity between the expert's family-specific coactivation profiles and their mean profile, and $\mathrm{Spec}_{l,e}$ is the maximum deviation from that mean profile across families.

For each layer, we select the top-scoring experts as generic experts. For each selected expert, we keep its TACG primary GPU and add a small number of secondary GPUs from structurally compatible non-primary GPUs, where compatibility is measured by coactivation affinity between the expert and the target card. This produces
\begin{equation}
    \mathcal{C}_{l,e}
    =
    \{g_{l,e}^{\mathrm{primary}}\}
    \cup
    \{g_{l,e}^{(1)},g_{l,e}^{(2)},\ldots\}.
    \label{eq:candidate-set}
\end{equation}
For non-selected experts, the candidate set contains only the primary GPU. Because the secondary candidates are both sparse and coactivation-compatible with TACG, GESR preserves TACG's communication structure while giving generic experts limited routing flexibility.

\paragraph{Locality- and load-aware serving.}
At inference time, the MoE router output is left unchanged; the serving system only chooses which candidate GPU serves a selected expert. Let $\mathcal{R}$ be the set of GPUs already accessed by the current token, $a$ the token's anchor GPU, and $f(x)$ its task family. We maintain an exponentially decayed recent load $L_m$ for each GPU and define
\begin{equation}
    \bar{L} = \frac{1}{M}\sum_m L_m .
    \label{eq:mean-load}
\end{equation}
For a generic expert, candidates whose recent load is within a tolerance $\theta$ form the feasible set
\begin{equation}
    \mathcal{C}_{l,e}^{\mathrm{feas}}
    =
    \{m\in\mathcal{C}_{l,e}: L_m \le (1+\theta)\bar{L}\}.
    \label{eq:feasible-set}
\end{equation}
If the feasible set is empty, we fall back to the full candidate set $\mathcal{C}_{l,e}$. After dispatch, the recent load is updated as
\begin{equation}
    L_m \leftarrow \rho L_m + \Delta_m ,
    \label{eq:load-update}
\end{equation}
where $\rho\in(0,1)$ is the decay factor and $\Delta_m$ is the dispatch increment contributed by the current token.

\section{Experiments}
\label{sec:experiments}

In this section, we conduct experiments to address the following research questions:
\begin{itemize}[leftmargin=*]
\item \textbf{RQ1}: Does TACG + GESR reduce communication more than strong placement baselines while preserving load balance?
\item \textbf{RQ2}: Does the advantage of TACG+GESR reflect genuine task-aware decoupling rather than a calibration artifact, and does it persist under skewed, unseen task distributions where same-family experts remain co-located and GESR absorbs residual hotspots?
\item \textbf{RQ3}: How much does each design choice—task-modulated coactivation, task-family preference, generic-expert replication, and the load guard—contribute to the final performance?
\item \textbf{RQ4}: Is GESR's specificity-penalized replica-selection criterion the right one, and does it remain effective under task skew?
\end{itemize}

\subsection{Experimental Setup}
\label{sec:exp_setup}
Following Section~\ref{sec:primary_placement}, we partition multi-task traffic into $K{=}4$ task families: \texttt{code}, \texttt{query}, \texttt{math}, and \texttt{complex\_reasoning}, each populated from widely used open datasets (full list in Appendix~\ref{app:exp_setup}). Calibration traces feed per-family usage and coactivation statistics, but are never reused as evaluation streams. We deploy each model on a $16$-GPU expert-parallel target ($M{=}16$), allocating $4$ GPUs to each of the four task families. Per-GPU capacities saturate the per-family budgets ($\sum_m C_{f,m}{=}B_f$ and $\sum_f B_f{=}E$). All methods are evaluated under identical capacity constraints.

\paragraph{Models.}

We evaluate our framework on three representative open-source MoE language models that span two architectural styles and three vendors: \textbf{DeepSeek-MoE-16B}~\citep{dai-etal-2024-deepseekmoe}, \textbf{Qwen1.5-MoE-A2.7B}~\citep{bai2023qwentechnicalreport}, and \textbf{Moonlight-16B-A3B}~\citep{liu2025muonscalablellmtraining}.

\paragraph{Baselines.}
We compare against a set of strong recent placement baselines: (i) \textbf{Baseline}, the model's default expert placement; (ii) \textbf{Frequency} and (iii) \textbf{Coactivation}, two canonical statistical placements; (iv) \textbf{Sem.P.}~\citep{li2026semantic}, a semantic-parallelism placement; (v) \textbf{ExFlow}~\citep{10579139}; (vi) \textbf{MoETuner}~\citep{go2025moetuneroptimizedmixtureexpert}; and (vii) \textbf{C2R}~\citep{zhang-etal-2025-advancing}. All baselines use the same experimental setup. Further models and baselines details are provided in Appendix~\ref{app:model_intro},~\ref{app:baseline_methods}.


\paragraph{Metrics.}
We report four system-level metrics that jointly characterize communication efficiency and load balance: \textbf{Comm}, \textbf{Comm.~Red.}, \textbf{Jain} and \textbf{MaxVio}. A detailed explanation is provided in Appendix~\ref{app:metric_intro}.



\subsection{Main Comparison with Placement Baselines (RQ1)}
\label{sec:main_results}

Table~\ref{tab:main_comm_jain} summarizes the primary comparison on the canonical multi-task workload across all three MoE models. Our framework (TACG\,+\,GESR) achieves the highest communication reduction on every model and on the cross-model average, while simultaneously attaining a load-balance profile that is strictly superior to every non-trivial baseline.

\begin{table*}[h]
\centering
\small
\setlength{\tabcolsep}{4.5pt}
\renewcommand{\arraystretch}{1.15}
\caption{Main comparison on three MoE models under the canonical multi-task workload.
For \textbf{Comm} and \textbf{MaxVio}, lower is better.
For \textbf{Comm. Red.} and \textbf{Jain}, higher is better.
The best result in each row is \textbf{bolded}, and the second best is \underline{underlined}.
\textbf{Ours} refers to our full framework.}
\label{tab:main_comm_jain}
\resizebox{\textwidth}{!}{
\begin{tabular}{ccc|ccccccc}
\toprule
\multirow{2}{*}[-0.3em]{\textbf{Model}}
& \multirow{2}{*}[-0.3em]{\textbf{Metric}}
& \multirow{2}{*}[-0.3em]{\textbf{Baseline}}
& \multicolumn{7}{c}{\textbf{Methods}} \\
\cmidrule(lr){4-10}
& & & \textbf{Freq.} & \textbf{Coact.} & \textbf{Sem.P.} & \textbf{ExFlow} & \textbf{MoETuner} & \textbf{C2R} & \textbf{Ours} \\
\midrule

\multirow{4}{*}{DeepSeek-MoE-16B}
& Comm $\downarrow$
& 112.3868 & 106.3577 & 99.0645 & 90.3942 & 90.6355 & \underline{87.9160} & 93.0491 & \textbf{76.1079} \\
& Comm. Red. $\uparrow$
& 0.00\% & 5.36\% & 11.85\% & 19.57\% & 19.35\% & \underline{21.77\%} & 17.21\% & \textbf{32.28\%} \\
& Jain $\uparrow$
& 0.9988 & 0.9345 & 0.9830 & 0.9962 & 0.9795 & \underline{0.9981} & 0.9900 & \textbf{0.9988} \\
& MaxVio $\downarrow$
& 0.0693 & 0.5765 & 0.2282 & 0.1753 & 0.3050 & \underline{0.0627} & 0.2575 & \textbf{0.0472} \\
\midrule

\multirow{4}{*}{Qwen1.5-MoE-A2.7B}
& Comm $\downarrow$
& 65.1870 & 62.1471 & 58.4267 & 53.9754 & \underline{53.4238} & 54.2211 & 55.3597 & \textbf{44.6977} \\
& Comm. Red. $\uparrow$
& 0.00\% & 4.66\% & 10.37\% & 17.20\% & \underline{18.05\%} & 16.82\% & 15.08\% & \textbf{31.43\%} \\
& Jain $\uparrow$
& 0.9877 & 0.9277 & 0.9585 & 0.9890 & 0.9662 & \textbf{0.9993} & 0.9855 & \underline{0.9991} \\
& MaxVio $\downarrow$
& 0.1147 & 0.4856 & 0.3141 & 0.1392 & 0.3995 & \underline{0.0605} & 0.1885 & \textbf{0.0596} \\
\midrule

\multirow{4}{*}{Moonlight-16B-A3B}
& Comm $\downarrow$
& 107.9992 & 93.2849 & 97.3618 & 94.2010 & \underline{87.4309} & 97.5114 & 94.7745 & \textbf{75.0888} \\
& Comm. Red. $\uparrow$
& 0.00\% & 13.62\% & 9.85\% & 12.78\% & \underline{19.04\%} & 9.71\% & 12.25\% & \textbf{30.47\%} \\
& Jain $\uparrow$
& 0.9894 & 0.7186 & 0.8803 & 0.9814 & 0.7263 & \underline{0.9943} & 0.9249 & \textbf{0.9947} \\
& MaxVio $\downarrow$
& 0.2002 & 1.8053 & 0.6197 & 0.3814 & 2.0368 & \underline{0.1791} & 0.7825 & \textbf{0.1141} \\
\midrule

\multirow{4}{*}{Across Models (Avg.)}
& Comm $\downarrow$
& 95.1910 & 87.2632 & 84.9510 & 79.5235 & \underline{77.1634} & 79.8829 & 81.0611 & \textbf{65.2981} \\
& Comm. Red. $\uparrow$
& 0.00\% & 7.88\% & 10.69\% & 16.51\% & \underline{18.81\%} & 16.10\% & 14.84\% & \textbf{31.39\%} \\
& Jain $\uparrow$
& 0.9920 & 0.8603 & 0.9406 & 0.9889 & 0.8907 & \underline{0.9972} & 0.9668 & \textbf{0.9975} \\
& MaxVio $\downarrow$
& 0.1281 & 0.9558 & 0.3873 & 0.2320 & 0.9137 & \underline{0.1008} & 0.4095 & \textbf{0.0736} \\
\bottomrule
\end{tabular}
}
\end{table*}

\paragraph{Obs.\,1: Task-aware placement yields the largest communication reduction by a substantial margin.}
Our method reduces per-token communication by \textbf{31.39\%} on average, \textbf{12.58\%} better than the strongest prior (ExFlow, 18.81\%) and \textbf{14.88\%--23.51\%} better than the other five baselines. The reduction is consistent across models (\textbf{30.47\%--32.28\%}), confirming the gain stems from the placement principle rather than any specific router or expert count.

\paragraph{Obs.\,2: The communication reduction is not obtained at the cost of load balance.}
Prior methods often sacrifice load balance. e.g., Frequency and ExFlow cause MaxVio $>$ 1 and Jain $<$ 0.9 on Moonlight, while Coactivation, Semantic Parallelism, and C2R inflate MaxVio on multiple models. In contrast, our framework achieves the \textbf{best MaxVio} on every model (\textbf{0.0472--0.1141}); Table~\ref{tab:main_comm_jain}) and an average Jain of \textbf{0.9975}, matching the Baseline's fairness while \textbf{cutting communication by 1/3}. It also uniformly outperforms MoETuner, the most load-balance-oriented prior.

\subsection{Is the Gain a Genuine Task-Aware Decoupling? (RQ2)}
\label{sec:rq2_taskaware}

\begin{figure}[h]
    \centering
    \begin{subfigure}[b]{0.48\linewidth}
        \centering
        \includegraphics[width=\linewidth]{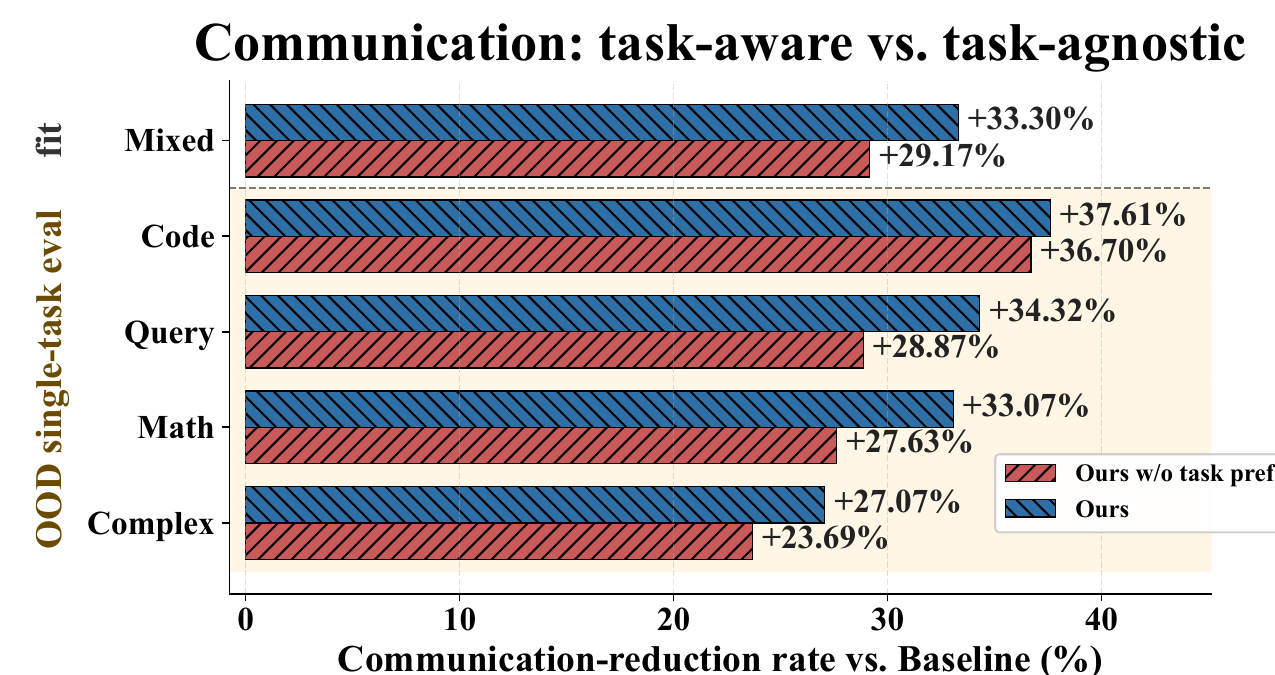}
        \caption{Communication under task skew.}
        \label{fig:skew_comm}
    \end{subfigure}
    \hfill
    \begin{subfigure}[b]{0.48\linewidth}
        \centering
        \includegraphics[width=\linewidth]{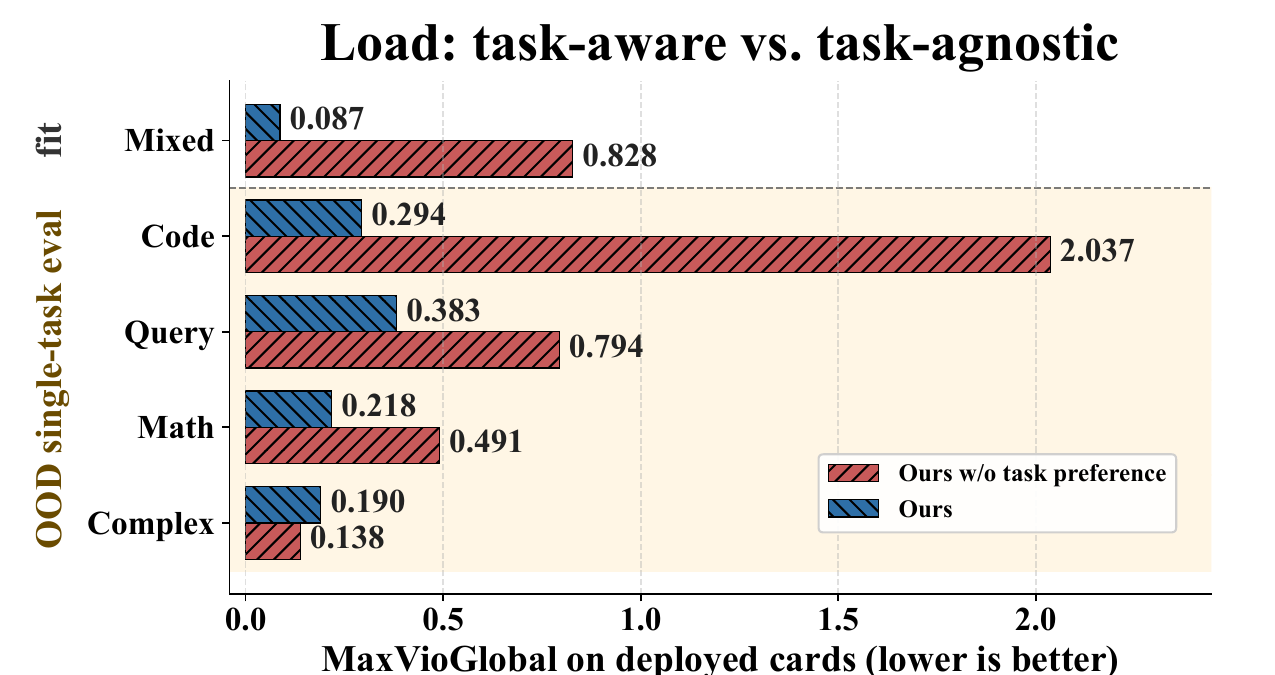}
        \caption{MaxVio on deployed cards under task skew.}
        \label{fig:skew_maxvio}
    \end{subfigure}
    \caption{Robustness under task-distribution skew on DeepSeek-MoE-16B.}
    \label{fig:skew}
\end{figure}

Table~\ref{tab:main_comm_jain} establishes a quantitative advantage of our framework on the canonical mix, but does not, by itself, adjudicate its underlying cause. A possible alternative explanation is that TACG has overfit to a one-shot statistic of the calibration mix, rather than recovering task-conditioned structure that generalizes beyond it. A genuine task-aware decoupling should therefore remain effective when the online task distribution deviates substantially from the calibration distribution. We test this hypothesis by holding the placement fixed at its canonical calibration and evaluating it under a sweep of workload-skew regimes to which it has not been fitted (Figure~\ref{fig:skew}, DeepSeek-MoE-16B).

\paragraph{Obs. 3: Communication advantage persists under severe task-distribution skew.}

Since the two placements in Figure~\ref{fig:skew} differ only in task-family preference, the gap between \textbf{Ours} and \textbf{Ours\,w/o\,task preference} isolates task-awareness. Under OOD single-task traffic, removing task-awareness causes MaxVioGlobal to surge---2.037 vs.\ 0.294 on \texttt{Code} ($\approx 7\times$), and $\approx 2\times$ on \texttt{Math} and \texttt{Query} (Fig.~\ref{fig:skew_maxvio})---while \textbf{Ours} remains low. Task-aware alignment lets a single-task burst spread across the expert family's GPU set; a task-agnostic graph instead packs heterogeneous families onto the same devices and collapses the burst into a hotspot. As Fig.~\ref{fig:skew_comm} shows, \textbf{Ours} achieves a markedly steadier reduction rate with far lower variance across workloads, whereas the task-agnostic variant suffers from large swings. Such stable, low-variance gains confirm that task-aware placement strictly advances the communication--balance frontier, not a one-shot calibration artifact.

\subsection{Ablation and Design-Choice Analysis (RQ3)}
\label{sec:ablation}

We next decompose the contribution of each building block of our framework. Cross-model averages are reported in Table~\ref{tab:ablation}.

\begin{table}[h]
\centering
\small
\setlength{\tabcolsep}{6pt}
\renewcommand{\arraystretch}{1.1}
\caption{Ablation study results averaged over three models. }
\label{tab:ablation}
\begin{tabular}{lcccc}
\toprule
\textbf{Variant} & \textbf{Avg Comm. Red.} $\uparrow$ & $\Delta$\textbf{Comm} & \textbf{Avg Jain} $\uparrow$ & \textbf{Avg MaxVio} $\downarrow$ \\
\midrule
Ours (TACG + GESR)        & \textbf{+31.39\%} & $-$      & \textbf{0.9975} & \textbf{0.0736} \\
\quad w/o coactivation    & +16.51\%          & $-14.88$ & 0.9955          & 0.0921 \\
\quad w/o task preference & +21.41\%          & $-9.98$  & 0.9537          & 0.2416 \\
\quad w/o GESR            & +24.23\%          & $-7.16$  & 0.9825          & 0.2763 \\
\quad w/o load guard      & +35.87\%          & $+4.48$  & 0.9895          & 0.1422 \\
\bottomrule
\end{tabular}
\end{table}

\paragraph{Obs.\,4: Task-conditioned coactivation dominates; both signals are necessary only jointly.}
Replacing the task-modulated coactivation graph with a uniform affinity (\emph{w/o coactivation}) drops the communication reduction from $31.39\%$ to $16.51\%$; disabling task-family preference ($\alpha{=}0$) reduces it by $9.98\%$ and triples MaxVio ($0.0736{\to}0.2416$). Only their joint interaction recovers the full $+31.39\%$ gain, confirming that task-conditioning is indispensable and the two signals are complementary.

\paragraph{Obs.\,5: GESR acts as a targeted load-balance safeguard.}
Removing GESR keeps communication gains high ($+24.23\%$) but nearly quadruples 
MaxVio (0.0736$\to$0.2763) and pushes the Jain index below 0.99. 
With GESR but $\theta$ disabled, reduction surges to $+35.87\%$ at the cost of 
MaxVio~0.1422. Only the full TACG\,+\,GESR 
with active $\theta$ reaches the communication--balance frontier, thereby 
separating the framework into a primary communication-reduction module 
(task-conditioned coactivation) and a load-balance safeguard (GESR with $\theta$), 
as designed in Section~\ref{sec:exp_setup}.

\subsection{GESR Replica-Selection Design under Severe Task Skew (RQ4)}
\label{sec:rq4_gesr_design}

Obs.~5 shows that removing GESR inflates MaxVio, but does not yet justify its specific form---centrality plus cross-family consistency reward and family-specificity penalty (Section~\ref{sec:gesr}). We therefore isolate the replica-selection criterion and ask whether GESR's specificity-aware design pays off under task-distribution drift. We reuse the extreme single-family skew setup of Section~\ref{sec:rq2_taskaware}: TACG's primary placement is held fixed at its canonical calibration, and we vary only the replica scorer across \textsc{NoRep}, \textsc{Freq-TopK}, \textsc{Hot}, \textsc{Coact-Cent} (centrality-only, a stripped-down GESR), \textsc{Random}, and \textsc{Gesr}.



\begin{figure}[h]
    \centering
    \begin{subfigure}[b]{0.48\linewidth}
        \centering
        \includegraphics[width=\linewidth]{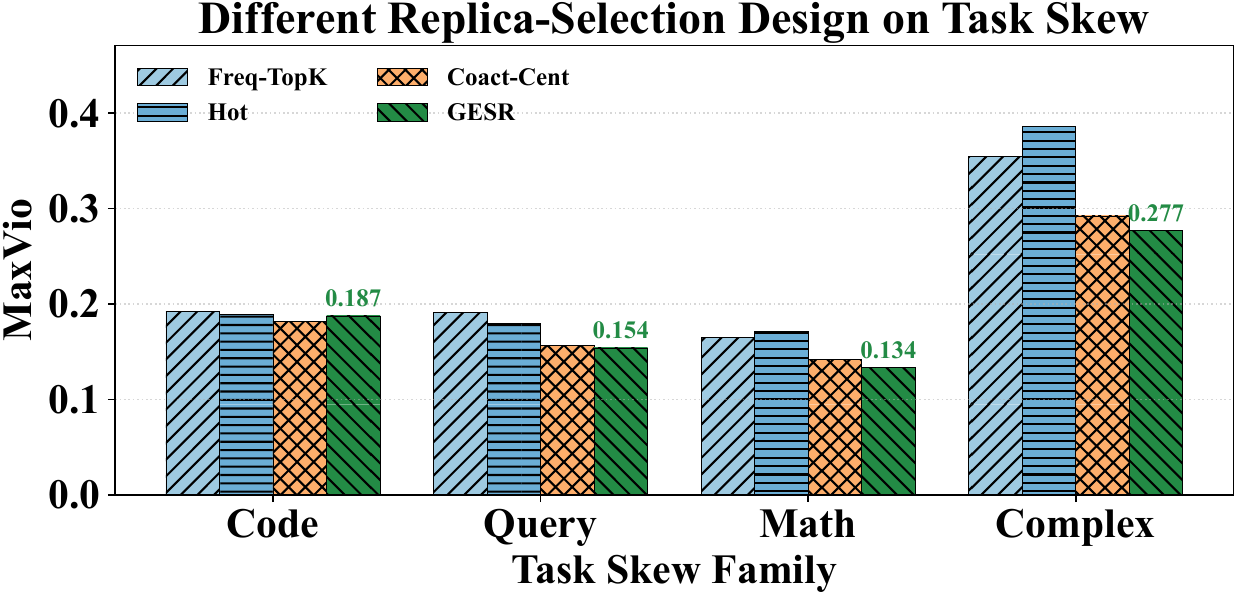}
        \caption{MaxVio on Task Skew (lower is better).}
        \label{fig:lofo_maxvio}
    \end{subfigure}
    \hfill
    \begin{subfigure}[b]{0.48\linewidth}
        \centering
        \includegraphics[width=\linewidth]{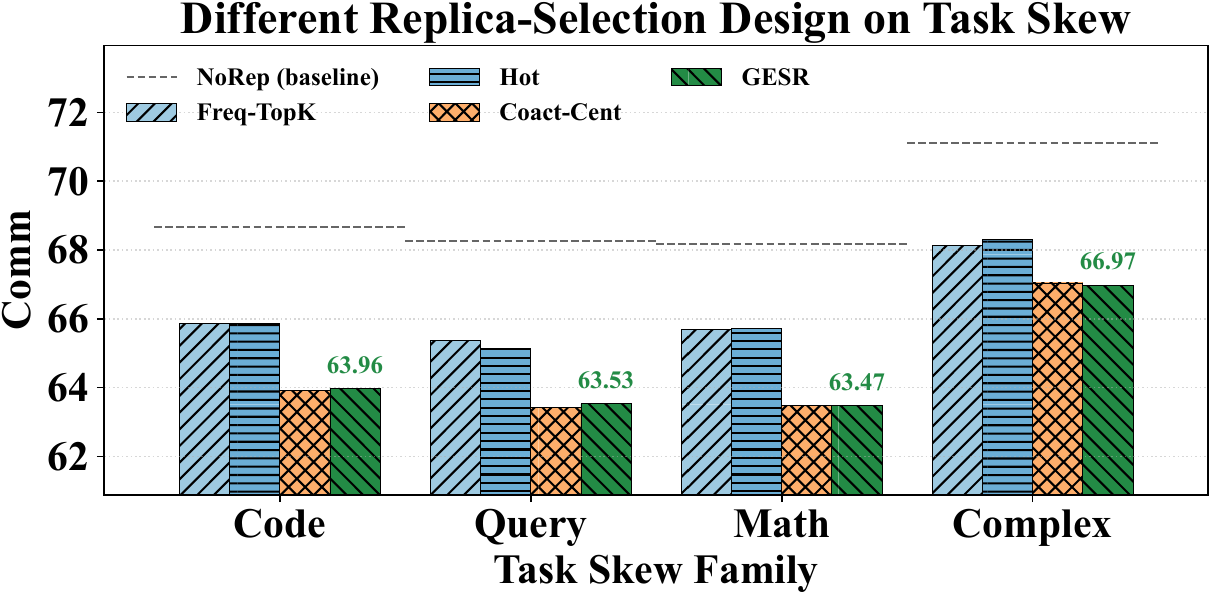}
        \caption{Per-token Communication (lower is better).}
        \label{fig:lofo_comm}
    \end{subfigure}
    \caption{Replica-selection Comparison on Task Skew.}
    \label{fig:lofo}
\end{figure}

\paragraph{Obs.\,6: GESR's specificity-aware scorer improves hotspot robustness at no communication cost.}
\textsc{Gesr} attains the lowest MaxVio on unseen families (Figure~\ref{fig:lofo_maxvio}), reducing hotspots by $2.1\%$ over the best coactivation-only baseline and by $15\text{--}17\%$ over frequency/hot baselines, while matching \textsc{Coact-Cent} on communication (Figure~\ref{fig:lofo_comm}). The gain is therefore a strict improvement.

\section{Limitations and Future Work}
\label{sec:limit}
Our work focuses on a \emph{static} offline placement and leaves three directions open. 
\emph{(i) Dynamic regrouping:} we recompute placement only on substantial workload drift; online regrouping within a request window could further close the gap to oracle placement under rapid shifts.
\emph{(ii) Memory co-design:} we enforce a hard per-GPU memory bound; integrating expert prefetching, caching, and offloading with our placement is a promising orthogonal direction for additional latency gains.
\emph{(iii) Specialized experts:} our evaluation targets general-purpose MoE models; we expect larger improvements from TACG and GESR on models with explicitly specialized experts (e.g., domain- or modality-partitioned), and we leave a systematic study for future work.

\section{Conclusion}
We show that expert coactivation in multi-task MoE serving is strongly task-conditioned and should directly drive placement.
TACG exploits this signal to build a capacity-feasible primary placement, and GESR replicates a few generic experts with load-aware online selection to protect locality under drift.
Across three MoE models, the combined framework cuts per-token communication by $31.39\%$ while keeping Jain $0.9975$ and MaxVio $0.0736$ (Tables~\ref{tab:main_comm_jain},~\ref{tab:ablation}).

\bibliographystyle{unsrtnat}
\bibliography{reference}

@inproceedings{
lepikhin2021gshard,
title={{\{}GS{\}}hard: Scaling Giant Models with Conditional Computation and Automatic Sharding},
author={Dmitry Lepikhin and HyoukJoong Lee and Yuanzhong Xu and Dehao Chen and Orhan Firat and Yanping Huang and Maxim Krikun and Noam Shazeer and Zhifeng Chen},
booktitle={International Conference on Learning Representations},
year={2021},
url={https://openreview.net/forum?id=qrwe7XHTmYb}
}

@article{JMLR:v23:21-0998,
  author  = {William Fedus and Barret Zoph and Noam Shazeer},
  title   = {Switch Transformers: Scaling to Trillion Parameter Models with Simple and Efficient Sparsity},
  journal = {Journal of Machine Learning Research},
  year    = {2022},
  volume  = {23},
  number  = {120},
  pages   = {1--39},
  url     = {http://jmlr.org/papers/v23/21-0998.html}
}

@inproceedings{
zhou2022mixtureofexperts,
title={Mixture-of-Experts with Expert Choice Routing},
author={Yanqi Zhou and Tao Lei and Hanxiao Liu and Nan Du and Yanping Huang and Vincent Y Zhao and Andrew M. Dai and Zhifeng Chen and Quoc V Le and James Laudon},
booktitle={Advances in Neural Information Processing Systems},
editor={Alice H. Oh and Alekh Agarwal and Danielle Belgrave and Kyunghyun Cho},
year={2022},
url={https://openreview.net/forum?id=jdJo1HIVinI}
}

@InProceedings{pmlr-v139-lewis21a,
  title = 	 {BASE Layers: Simplifying Training of Large, Sparse Models},
  author =       {Lewis, Mike and Bhosale, Shruti and Dettmers, Tim and Goyal, Naman and Zettlemoyer, Luke},
  booktitle = 	 {Proceedings of the 38th International Conference on Machine Learning},
  pages = 	 {6265--6274},
  year = 	 {2021},
  editor = 	 {Meila, Marina and Zhang, Tong},
  volume = 	 {139},
  series = 	 {Proceedings of Machine Learning Research},
  month = 	 {18--24 Jul},
  publisher =    {PMLR},
  url = 	 {https://proceedings.mlr.press/v139/lewis21a.html},
}

@inproceedings{dai-etal-2024-deepseekmoe,
    title = "{D}eep{S}eek{M}o{E}: Towards Ultimate Expert Specialization in Mixture-of-Experts Language Models",
    author = "Dai, Damai  and
      Deng, Chengqi  and
      Zhao, Chenggang  and
      Xu, R.x.  and
      Gao, Huazuo  and
      Chen, Deli  and
      Li, Jiashi  and
      Zeng, Wangding  and
      Yu, Xingkai  and
      Wu, Y.  and
      Xie, Zhenda  and
      Li, Y.k.  and
      Huang, Panpan  and
      Luo, Fuli  and
      Ruan, Chong  and
      Sui, Zhifang  and
      Liang, Wenfeng",
    editor = "Ku, Lun-Wei  and
      Martins, Andre  and
      Srikumar, Vivek",
    booktitle = "Proceedings of the 62nd Annual Meeting of the Association for Computational Linguistics (Volume 1: Long Papers)",
    month = aug,
    year = "2024",
    address = "Bangkok, Thailand",
    publisher = "Association for Computational Linguistics",
    url = "https://aclanthology.org/2024.acl-long.70/",
    doi = "10.18653/v1/2024.acl-long.70",
    pages = "1280--1297"
}

@inproceedings{MLSYS2023_5616d34c,
 author = {Hwang, Changho and Cui, Wei and Xiong, Yifan and Yang, Ziyue and Liu, Ze and Hu, Han and Wang, Zilong and Salas, Rafael and Jose, Jithin and Ram, Prabhat and Chau, HoYuen and Cheng, Peng and Yang, Fan and Yang, Mao and Xiong, Yongqiang},
 booktitle = {Proceedings of Machine Learning and Systems},
 editor = {D. Song and M. Carbin and T. Chen},
 pages = {269--287},
 publisher = {Curan},
 title = {Tutel: Adaptive Mixture-of-Experts at Scale},
 url = {https://proceedings.mlsys.org/paper_files/paper/2023/file/5616d34cf8ff73942cfd5aa922842556-Paper-mlsys2023.pdf},
 volume = {5},
 year = {2023}
}

@inproceedings{MLSYS2023_5a54f793,
 author = {Gale, Trevor and Narayanan, Deepak and Young, Cliff and Zaharia, Matei},
 booktitle = {Proceedings of Machine Learning and Systems},
 editor = {D. Song and M. Carbin and T. Chen},
 pages = {288--304},
 publisher = {Curan},
 title = {MegaBlocks: Efficient Sparse Training with Mixture-of-Experts},
 url = {https://proceedings.mlsys.org/paper_files/paper/2023/file/5a54f79333768effe7e8927bcccffe40-Paper-mlsys2023.pdf},
 volume = {5},
 year = {2023}
}

@inproceedings{10.1145/3600006.3613165,
author = {Kwon, Woosuk and Li, Zhuohan and Zhuang, Siyuan and Sheng, Ying and Zheng, Lianmin and Yu, Cody Hao and Gonzalez, Joseph and Zhang, Hao and Stoica, Ion},
title = {Efficient Memory Management for Large Language Model Serving with PagedAttention},
year = {2023},
isbn = {9798400702297},
publisher = {Association for Computing Machinery},
address = {New York, NY, USA},
url = {https://doi.org/10.1145/3600006.3613165},
doi = {10.1145/3600006.3613165},
booktitle = {Proceedings of the 29th Symposium on Operating Systems Principles},
pages = {611–626},
numpages = {16},
location = {Koblenz, Germany},
series = {SOSP '23}
}

@inproceedings {288705,
author = {Jiamin Li and Yimin Jiang and Yibo Zhu and Cong Wang and Hong Xu},
title = {Accelerating Distributed {MoE} Training and Inference with Lina},
booktitle = {2023 USENIX Annual Technical Conference (USENIX ATC 23)},
year = {2023},
isbn = {978-1-939133-35-9},
address = {Boston, MA},
pages = {945--959},
url = {https://www.usenix.org/conference/atc23/presentation/li-jiamin},
publisher = {USENIX Association},
month = jul
}

@INPROCEEDINGS{10579139,
  author={Yao, Jinghan and Anthony, Quentin and Shafi, Aamir and Subramoni, Hari and DK Panda, Dhabaleswar K.},
  booktitle={2024 IEEE International Parallel and Distributed Processing Symposium (IPDPS)}, 
  title={Exploiting Inter-Layer Expert Affinity for Accelerating Mixture-of-Experts Model Inference}, 
  year={2024},
  volume={},
  number={},
  pages={915-925},
  doi={10.1109/IPDPS57955.2024.00086}}

@inproceedings{zhang-etal-2025-advancing,
    title = "Advancing {M}o{E} Efficiency: A Collaboration-Constrained Routing ($\texttt{C2R}$) Strategy for Better Expert Parallelism Design",
    author = "Zhang, Mohan  and
      Li, Pingzhi  and
      Peng, Jie  and
      Qiu, Mufan  and
      Chen, Tianlong",
    editor = "Chiruzzo, Luis  and
      Ritter, Alan  and
      Wang, Lu",
    booktitle = "Proceedings of the 2025 Conference of the Nations of the Americas Chapter of the Association for Computational Linguistics: Human Language Technologies (Volume 1: Long Papers)",
    month = apr,
    year = "2025",
    address = "Albuquerque, New Mexico",
    publisher = "Association for Computational Linguistics",
    url = "https://aclanthology.org/2025.naacl-long.347/",
    doi = "10.18653/v1/2025.naacl-long.347",
    pages = "6815--6825",
    ISBN = "979-8-89176-189-6"
}

@misc{go2025moetuneroptimizedmixtureexpert,
      title={MoETuner: Optimized Mixture of Expert Serving with Balanced Expert Placement and Token Routing}, 
      author={Seokjin Go and Divya Mahajan},
      year={2025},
      eprint={2502.06643},
      archivePrefix={arXiv},
      primaryClass={cs.LG},
      url={https://arxiv.org/abs/2502.06643}, 
}

@misc{han2026gracemoegroupingreplicationlocalityaware,
      title={GRACE-MoE: Grouping and Replication with Locality-Aware Routing for Efficient Distributed MoE Inference}, 
      author={Yu Han and Lehan Pan and Jie Peng and Ziyang Tao and Wuyang Zhang and Yanyong Zhang},
      year={2026},
      eprint={2509.25041},
      archivePrefix={arXiv},
      primaryClass={cs.DC},
      url={https://arxiv.org/abs/2509.25041}, 
}

@inproceedings{10.1145/3219819.3220007,
author = {Ma, Jiaqi and Zhao, Zhe and Yi, Xinyang and Chen, Jilin and Hong, Lichan and Chi, Ed H.},
title = {Modeling Task Relationships in Multi-task Learning with Multi-gate Mixture-of-Experts},
year = {2018},
isbn = {9781450355520},
publisher = {Association for Computing Machinery},
address = {New York, NY, USA},
url = {https://doi.org/10.1145/3219819.3220007},
doi = {10.1145/3219819.3220007},
booktitle = {Proceedings of the 24th ACM SIGKDD International Conference on Knowledge Discovery \& Data Mining},
pages = {1930–1939},
numpages = {10},
location = {London, United Kingdom},
series = {KDD '18}
}

@inproceedings{10.5555/3540261.3542507,
author = {Hazimeh, Hussein and Zhao, Zhe and Chowdhery, Aakanksha and Sathiamoorthy, Maheswaran and Chen, Yihua and Mazumder, Rahul and Hong, Lichan and Chi, Ed H.},
title = {DSelect-k: differentiable selection in the mixture of experts with applications to multi-task learning},
year = {2021},
isbn = {9781713845393},
publisher = {Curran Associates Inc.},
address = {Red Hook, NY, USA},
booktitle = {Proceedings of the 35th International Conference on Neural Information Processing Systems},
articleno = {2246},
numpages = {13},
series = {NIPS '21}
}

@inproceedings{kudugunta-etal-2021-beyond-distillation,
    title = "Beyond Distillation: Task-level Mixture-of-Experts for Efficient Inference",
    author = "Kudugunta, Sneha  and
      Huang, Yanping  and
      Bapna, Ankur  and
      Krikun, Maxim  and
      Lepikhin, Dmitry  and
      Luong, Minh-Thang  and
      Firat, Orhan",
    editor = "Moens, Marie-Francine  and
      Huang, Xuanjing  and
      Specia, Lucia  and
      Yih, Scott Wen-tau",
    booktitle = "Findings of the Association for Computational Linguistics: EMNLP 2021",
    month = nov,
    year = "2021",
    address = "Punta Cana, Dominican Republic",
    publisher = "Association for Computational Linguistics",
    url = "https://aclanthology.org/2021.findings-emnlp.304/",
    doi = "10.18653/v1/2021.findings-emnlp.304",
    pages = "3577--3599"
}

@misc{chen2022taskspecificexpertpruningsparse,
      title={Task-Specific Expert Pruning for Sparse Mixture-of-Experts}, 
      author={Tianyu Chen and Shaohan Huang and Yuan Xie and Binxing Jiao and Daxin Jiang and Haoyi Zhou and Jianxin Li and Furu Wei},
      year={2022},
      eprint={2206.00277},
      archivePrefix={arXiv},
      primaryClass={cs.LG},
      url={https://arxiv.org/abs/2206.00277}, 
}

@inproceedings{ICLR2025_e43a3399,
 author = {Liao, Mengqi and Chen, Wei and Shen, Junfeng and Guo, Shengnan and Wan, Huaiyu},
 booktitle = {International Conference on Learning Representations},
 editor = {Y. Yue and A. Garg and N. Peng and F. Sha and R. Yu},
 pages = {91309--91330},
 title = {HMoRA: Making LLMs More Effective with Hierarchical Mixture of LoRA Experts},
 url = {https://proceedings.iclr.cc/paper_files/paper/2025/file/e43a33994a28f746dcfd53eb51ed3c2d-Paper-Conference.pdf},
 volume = {2025},
 year = {2025}
}

@inproceedings{pham-etal-2023-task,
    title = "Task-Based {M}o{E} for Multitask Multilingual Machine Translation",
    author = "Pham, Hai  and
      Kim, Young Jin  and
      Mukherjee, Subhabrata  and
      Woodruff, David P.  and
      Poczos, Barnabas  and
      Hassan, Hany",
    editor = "Ataman, Duygu",
    booktitle = "Proceedings of the 3rd Workshop on Multi-lingual Representation Learning (MRL)",
    month = dec,
    year = "2023",
    address = "Singapore",
    publisher = "Association for Computational Linguistics",
    url = "https://aclanthology.org/2023.mrl-1.13/",
    doi = "10.18653/v1/2023.mrl-1.13",
    pages = "164--172"
}

@article{shazeer2017outrageously,
  title={Outrageously large neural networks: The sparsely-gated mixture-of-experts layer},
  author={Shazeer, Noam and Mirhoseini, Azalia and Maziarz, Krzysztof and Davis, Andy and Le, Quoc and Hinton, Geoffrey and Dean, Jeff},
  journal={arXiv preprint arXiv:1701.06538},
  year={2017}
}

@article{fedus2022switch,
  title={Switch transformers: Scaling to trillion parameter models with simple and efficient sparsity},
  author={Fedus, William and Zoph, Barret and Shazeer, Noam},
  journal={Journal of Machine Learning Research},
  volume={23},
  number={120},
  pages={1--39},
  year={2022}
}

@article{gale2023megablocks,
  title={Megablocks: Efficient sparse training with mixture-of-experts},
  author={Gale, Trevor and Narayanan, Deepak and Young, Cliff and Zaharia, Matei},
  journal={Proceedings of Machine Learning and Systems},
  volume={5},
  pages={288--304},
  year={2023}
}

@article{hwang2023tutel,
  title={Tutel: Adaptive mixture-of-experts at scale},
  author={Hwang, Changho and Cui, Wei and Xiong, Yifan and Yang, Ziyue and Liu, Ze and Hu, Han and Wang, Zilong and Salas, Rafael and Jose, Jithin and Ram, Prabhat and others},
  journal={Proceedings of Machine Learning and Systems},
  volume={5},
  pages={269--287},
  year={2023}
}

@inproceedings{
li2026semantic,
title={Semantic Parallelism: Redefining Efficient MoE Inference via Model-Data Co-Scheduling},
author={Yan Li and Zhenyu Zhang and Zhengang Wang and Pengfei chen and Pengfei Zheng},
booktitle={The Fourteenth International Conference on Learning Representations},
year={2026},
url={https://openreview.net/forum?id=MSHPrMpIHZ}
}

@misc{bambhaniya2026scalingmultinodemixtureofexpertsinference,
      title={Scaling Multi-Node Mixture-of-Experts Inference Using Expert Activation Patterns}, 
      author={Abhimanyu Bambhaniya and Geonhwa Jeong and Jason Park and Jiecao Yu and Jaewon Lee and Pengchao Wang and Changkyu Kim and Chunqiang Tang and Tushar Krishna},
      year={2026},
      eprint={2604.23150},
      archivePrefix={arXiv},
      primaryClass={cs.LG},
      url={https://arxiv.org/abs/2604.23150}, 
}

@inproceedings{NEURIPS2025_4598de7d,
 author = {Guo, Hongcan and Lu, Haolang and Nan, Guoshun and Chu, Bolun and Zhuang, Jialin and Yang, Yuan and Che, Wenhao and Cao, Xinye and Leng, Sicong and Cui, Qimei and Jiang, Xudong},
 booktitle = {Advances in Neural Information Processing Systems},
 editor = {D. Belgrave and C. Zhang and H. Lin and R. Pascanu and P. Koniusz and M. Ghassemi and N. Chen},
 pages = {48767--48809},
 publisher = {Curran Associates, Inc.},
 title = {Advancing Expert Specialization for Better MoE},
 url = {https://proceedings.neurips.cc/paper_files/paper/2025/file/4598de7d243d528e38eb0c5d8155fb52-Paper-Conference.pdf},
 volume = {38},
 year = {2025}
}

@misc{austin2021programsynthesislargelanguage,
      title={Program Synthesis with Large Language Models}, 
      author={Jacob Austin and Augustus Odena and Maxwell Nye and Maarten Bosma and Henryk Michalewski and David Dohan and Ellen Jiang and Carrie Cai and Michael Terry and Quoc Le and Charles Sutton},
      year={2021},
      eprint={2108.07732},
      archivePrefix={arXiv},
      primaryClass={cs.PL},
      url={https://arxiv.org/abs/2108.07732}, 
}

@misc{yu2019spiderlargescalehumanlabeleddataset,
      title={Spider: A Large-Scale Human-Labeled Dataset for Complex and Cross-Domain Semantic Parsing and Text-to-SQL Task}, 
      author={Tao Yu and Rui Zhang and Kai Yang and Michihiro Yasunaga and Dongxu Wang and Zifan Li and James Ma and Irene Li and Qingning Yao and Shanelle Roman and Zilin Zhang and Dragomir Radev},
      year={2019},
      eprint={1809.08887},
      archivePrefix={arXiv},
      primaryClass={cs.CL},
      url={https://arxiv.org/abs/1809.08887}, 
}

@misc{cobbe2021trainingverifierssolvemath,
      title={Training Verifiers to Solve Math Word Problems}, 
      author={Karl Cobbe and Vineet Kosaraju and Mohammad Bavarian and Mark Chen and Heewoo Jun and Lukasz Kaiser and Matthias Plappert and Jerry Tworek and Jacob Hilton and Reiichiro Nakano and Christopher Hesse and John Schulman},
      year={2021},
      eprint={2110.14168},
      archivePrefix={arXiv},
      primaryClass={cs.LG},
      url={https://arxiv.org/abs/2110.14168}, 
}

@inproceedings{NEURIPS2023_89e44582,
 author = {Guha, Neel and Nyarko, Julian and Ho, Daniel and R\'{e}, Christopher and Chilton, Adam and K, Aditya and Chohlas-Wood, Alex and Peters, Austin and Waldon, Brandon and Rockmore, Daniel and Zambrano, Diego and Talisman, Dmitry and Hoque, Enam and Surani, Faiz and Fagan, Frank and Sarfaty, Galit and Dickinson, Gregory and Porat, Haggai and Hegland, Jason and Wu, Jessica and Nudell, Joe and Niklaus, Joel and Nay, John and Choi, Jonathan and Tobia, Kevin and Hagan, Margaret and Ma, Megan and Livermore, Michael and Rasumov-Rahe, Nikon and Holzenberger, Nils and Kolt, Noam and Henderson, Peter and Rehaag, Sean and Goel, Sharad and Gao, Shang and Williams, Spencer and Gandhi, Sunny and Zur, Tom and Iyer, Varun and Li, Zehua},
 booktitle = {Advances in Neural Information Processing Systems},
 editor = {A. Oh and T. Naumann and A. Globerson and K. Saenko and M. Hardt and S. Levine},
 pages = {44123--44279},
 publisher = {Curran Associates, Inc.},
 title = {LegalBench: A Collaboratively Built Benchmark for Measuring Legal Reasoning in Large Language Models},
 url = {https://proceedings.neurips.cc/paper_files/paper/2023/file/89e44582fd28ddfea1ea4dcb0ebbf4b0-Paper-Datasets_and_Benchmarks.pdf},
 volume = {36},
 year = {2023}
}

@misc{bai2023qwentechnicalreport,
      title={Qwen Technical Report}, 
      author={Jinze Bai and Shuai Bai and Yunfei Chu and Zeyu Cui and Kai Dang and Xiaodong Deng and Yang Fan and Wenbin Ge and Yu Han and Fei Huang and Binyuan Hui and Luo Ji and Mei Li and Junyang Lin and Runji Lin and Dayiheng Liu and Gao Liu and Chengqiang Lu and Keming Lu and Jianxin Ma and Rui Men and Xingzhang Ren and Xuancheng Ren and Chuanqi Tan and Sinan Tan and Jianhong Tu and Peng Wang and Shijie Wang and Wei Wang and Shengguang Wu and Benfeng Xu and Jin Xu and An Yang and Hao Yang and Jian Yang and Shusheng Yang and Yang Yao and Bowen Yu and Hongyi Yuan and Zheng Yuan and Jianwei Zhang and Xingxuan Zhang and Yichang Zhang and Zhenru Zhang and Chang Zhou and Jingren Zhou and Xiaohuan Zhou and Tianhang Zhu},
      year={2023},
      eprint={2309.16609},
      archivePrefix={arXiv},
      primaryClass={cs.CL},
      url={https://arxiv.org/abs/2309.16609}, 
}

@misc{liu2025muonscalablellmtraining,
      title={Muon is Scalable for LLM Training}, 
      author={Jingyuan Liu and Jianlin Su and Xingcheng Yao and Zhejun Jiang and Guokun Lai and Yulun Du and Yidao Qin and Weixin Xu and Enzhe Lu and Junjie Yan and Yanru Chen and Huabin Zheng and Yibo Liu and Shaowei Liu and Bohong Yin and Weiran He and Han Zhu and Yuzhi Wang and Jianzhou Wang and Mengnan Dong and Zheng Zhang and Yongsheng Kang and Hao Zhang and Xinran Xu and Yutao Zhang and Yuxin Wu and Xinyu Zhou and Zhilin Yang},
      year={2025},
      eprint={2502.16982},
      archivePrefix={arXiv},
      primaryClass={cs.LG},
      url={https://arxiv.org/abs/2502.16982}, 
}

@misc{chen2021evaluatinglargelanguagemodels,
      title={Evaluating Large Language Models Trained on Code}, 
      author={Mark Chen and Jerry Tworek and Heewoo Jun and Qiming Yuan and Henrique Ponde de Oliveira Pinto and Jared Kaplan and Harri Edwards and Yuri Burda and Nicholas Joseph and Greg Brockman and Alex Ray and Raul Puri and Gretchen Krueger and Michael Petrov and Heidy Khlaaf and Girish Sastry and Pamela Mishkin and Brooke Chan and Scott Gray and Nick Ryder and Mikhail Pavlov and Alethea Power and Lukasz Kaiser and Mohammad Bavarian and Clemens Winter and Philippe Tillet and Felipe Petroski Such and Dave Cummings and Matthias Plappert and Fotios Chantzis and Elizabeth Barnes and Ariel Herbert-Voss and William Hebgen Guss and Alex Nichol and Alex Paino and Nikolas Tezak and Jie Tang and Igor Babuschkin and Suchir Balaji and Shantanu Jain and William Saunders and Christopher Hesse and Andrew N. Carr and Jan Leike and Josh Achiam and Vedant Misra and Evan Morikawa and Alec Radford and Matthew Knight and Miles Brundage and Mira Murati and Katie Mayer and Peter Welinder and Bob McGrew and Dario Amodei and Sam McCandlish and Ilya Sutskever and Wojciech Zaremba},
      year={2021},
      eprint={2107.03374},
      archivePrefix={arXiv},
      primaryClass={cs.LG},
      url={https://arxiv.org/abs/2107.03374}, 
}

@misc{hendrycks2021measuringcodingchallengecompetence,
      title={Measuring Coding Challenge Competence With APPS}, 
      author={Dan Hendrycks and Steven Basart and Saurav Kadavath and Mantas Mazeika and Akul Arora and Ethan Guo and Collin Burns and Samir Puranik and Horace He and Dawn Song and Jacob Steinhardt},
      year={2021},
      eprint={2105.09938},
      archivePrefix={arXiv},
      primaryClass={cs.SE},
      url={https://arxiv.org/abs/2105.09938}, 
}

@inproceedings{LiNuminaMathTL,
  title={NuminaMath: The largest public dataset in AI4Maths with 860k pairs of competition math problems and solutions},
  author={Jia Li and Edward Beeching and Lewis Tunstall and Ben Lipkin and Roman Soletskyi and Shengyi Huang and Kashif Rasul and Long Yu and Albert Qiaochu Jiang and Ziju Shen and Zihan Qin and Bin Dong and Li Zhou and Yann Fleureau and Guillaume Lample and Stanislas Polu and Hugging Face and AI Mistral},
  url={https://api.semanticscholar.org/CorpusID:286088415}
}

@inproceedings{suzgun-etal-2023-challenging,
    title = "Challenging {BIG}-Bench Tasks and Whether Chain-of-Thought Can Solve Them",
    author = {Suzgun, Mirac  and
      Scales, Nathan  and
      Sch{\"a}rli, Nathanael  and
      Gehrmann, Sebastian  and
      Tay, Yi  and
      Chung, Hyung Won  and
      Chowdhery, Aakanksha  and
      Le, Quoc  and
      Chi, Ed  and
      Zhou, Denny  and
      Wei, Jason},
    editor = "Rogers, Anna  and
      Boyd-Graber, Jordan  and
      Okazaki, Naoaki",
    booktitle = "Findings of the Association for Computational Linguistics: ACL 2023",
    month = jul,
    year = "2023",
    address = "Toronto, Canada",
    publisher = "Association for Computational Linguistics",
    url = "https://aclanthology.org/2023.findings-acl.824/",
    doi = "10.18653/v1/2023.findings-acl.824",
    pages = "13003--13051"
}

@article{zhongSeq2SQL2017,
  author    = {Victor Zhong and
               Caiming Xiong and
               Richard Socher},
  title     = {Seq2SQL: Generating Structured Queries from Natural Language using
               Reinforcement Learning},
  journal   = {CoRR},
  volume    = {abs/1709.00103},
  year      = {2017}
}

\newpage
\appendix

\section{Sensitivity to the Calibration Token Budget}
\label{app:tokencap}

A practical concern for any calibration-based placement method is how many calibration tokens it actually needs. To answer this we run the full placement pipeline on DeepSeek-MoE-16B while sweeping the per-sample calibration cap from $1$ token to $2048$ tokens (with one calibration sample per family); under this protocol the resulting calibration corpus grows from $9$ tokens at \texttt{tokcap}$=1$ to $\sim\!2$k tokens at \texttt{tokcap}$=2048$. Six baselines and \textsc{Ours} are recalibrated on each budget and evaluated on the held-out test stream. All other settings (top-$k$, $M=16$, per-family GPU groups) follow the main protocol.

\begin{figure}[h]
\centering
\includegraphics[width=0.85\textwidth]{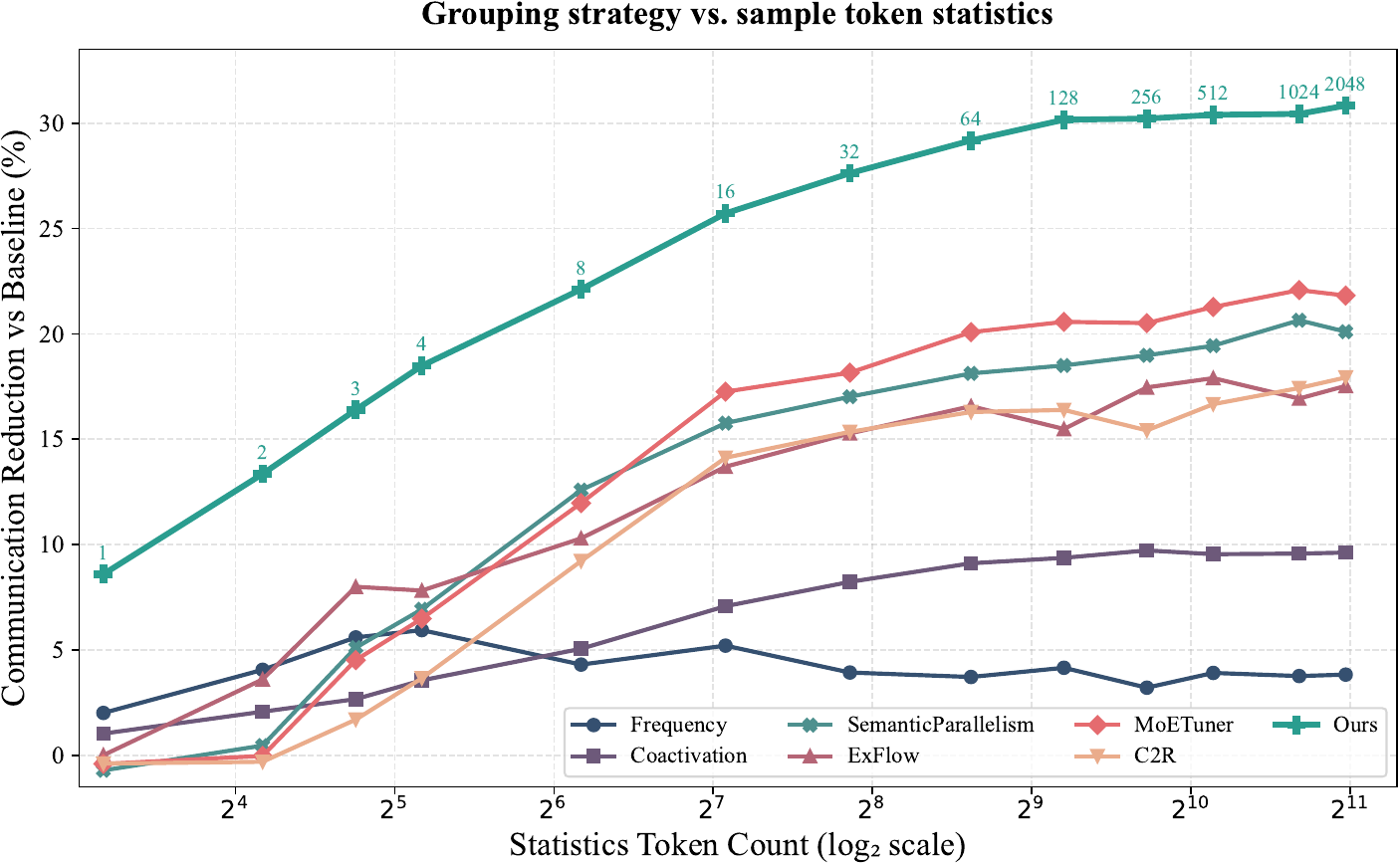}
\caption{Communication reduction vs.\ calibration token budget on DeepSeek-MoE-16B. The horizontal axis is the total number of calibration tokens used (log scale). \textsc{Ours} (TACG\,+\,GESR) attains $\sim\!18\%$ Comm.\ Red.\ already at $36$ tokens and saturates near $\sim\!30\%$ from $\sim\!600$ tokens onward, while pooled coactivation, frequency-only, and routing-pattern baselines plateau much lower.}
\label{fig:tokencap}
\end{figure}


\paragraph{Obs.\,A1: TACG\,+\,GESR converges with hundreds, not millions, of calibration tokens.}
With as few as $36$ calibration tokens \textsc{Ours} already exceeds $+18\%$ Comm.\ Red.\ on DeepSeek-MoE-16B, surpassing the full-budget Coactivation/Frequency/MoETuner/C2R baselines. The curve enters a plateau around $128$--$256$ tokens (Figure~\ref{fig:tokencap}); doubling the budget further from $\sim\!600$ to $\sim\!2{,}000$ tokens improves Comm.\ Red.\ by less than $0.7$ points. We attribute this fast convergence to the fact that TACG only needs to estimate per-family \emph{relative} coactivation orderings, not full distributions, and GESR's score $\mathrm{Cent}+\lambda_1\mathrm{Cons}-\lambda_2\mathrm{Spec}$ depends only on aggregate centrality and per-family means, both of which stabilize quickly.

\section{Hyperparameter Sensitivity}
\label{app:param_sweep}

Our framework exposes three hyperparameters that warrant a sensitivity study: (i) the TACG task-preference strength $\alpha$ that trades off pure co-activation locality against task-modulated grouping (Section~\ref{sec:primary_placement}); (ii) the GESR scoring coefficients $(\lambda_1,\lambda_2)$ that trade off cross-family consistency against family-specific deviation (Section~\ref{sec:gesr}); and (iii) the load-guard slack $\theta$ that bounds the secondary-card admission rule at serving time (Section~\ref{sec:gesr}). Each sweep varies one hyperparameter while holding all others at their canonical values ($\alpha=0.25$, $\lambda_1{=}0$, $\lambda_2{=}0$, $\theta=0.15$, replica budget $n_{\mathrm{rep}}{=}8$, $k_{\mathrm{sec}}{=}2$, decay $\rho{=}0.995$). Unless noted, every reported value is the cross-model average over DeepSeek-MoE-16B, Qwen1.5-MoE-A2.7B, and Moonlight-16B-A3B.

\subsection{Task-Preference Strength $\alpha$}
\label{app:sweep_alpha}

The task-preference strength $\alpha$ interpolates between task-agnostic co-activation grouping and TACG's task-modulated objective (Section~\ref{sec:primary_placement}). At $\alpha{=}0$ every expert is pinned to its physical home family, which recovers the task-agnostic family partition (equivalent to \emph{Ours w/o task preference} in the ablation table). We sweep $\alpha\!\in\!\{0, 0.05, 0.10, 0.25, 0.50, 0.75, 1.00\}$.

\begin{table}[h]
\centering
\caption{Sensitivity of \textsc{Ours} to the TACG task-preference strength $\alpha$, averaged over three MoE backbones. Once $\alpha\!\ge\!\alpha^{\star}{=}0.25$ the grouping is fully task-modulated and all metrics saturate; below the pivot the placement collapses toward a family-pinned partition and both communication and load balance degrade.}
\label{tab:sweep-alpha}
\begin{tabular}{cccccc}
\hline
$\alpha$ & Comm. Red. $\uparrow$ & Jain $\uparrow$ & MaxVio $\downarrow$ & CV $\downarrow$ & Comm $\downarrow$ \\
\hline
0.00 & +21.74\% & 0.9530 & 0.2510 & 0.1996 & 76.49 \\
0.05 & +21.71\% & 0.9566 & 0.2316 & 0.1930 & 76.60 \\
0.10 & +22.17\% & 0.9574 & 0.2189 & 0.1902 & 76.15 \\
0.25 & +31.85\% & 0.9980 & 0.0740 & 0.0418 & 66.89 \\
0.50 & +31.56\% & 0.9983 & 0.0787 & \textbf{0.0388} & 67.16 \\
0.75 & +31.79\% & 0.9979 & 0.0675 & 0.0400 & 66.90 \\
1.00 & +31.17\% & 0.9980 & 0.0765 & 0.0413 & 67.47 \\
\hline
\end{tabular}
\end{table}

\paragraph{Obs.\,A1: Task modulation saturates at $\alpha^{\star}$, and the canonical choice sits at the knee.}
Table~\ref{tab:sweep-alpha} shows a clear two-regime behaviour. In the \emph{family-pinned} regime ($\alpha<\alpha^{\star}$), Comm.\ Red.\ is stuck around $+21$--$22\%$ and MaxVio is an order of magnitude worse ($0.22$--$0.25$) because pinning experts to their physical home ignores the actual task-conditioned co-activation structure. Once $\alpha$ crosses the pivot, every metric improves abruptly: Comm.\ Red.\ jumps to $+31.85\%$ ($+10$\%), MaxVio drops to $0.074$ ($-0.18$), and Jain rises to $0.998$. Further increasing $\alpha$ up to $1.0$ changes Comm.\ Red.\ by at most $0.7$\% and MaxVio by $\le 0.013$, indicating that once TACG is activated the result is insensitive to the exact weight. We use $\alpha{=}0.25$ in the main paper because it is the smallest value that realises the full task-modulated benefit; the same conclusion holds on each individual backbone (DeepSeek: $+32.82\%$, Moonlight: $+30.93\%$, Qwen: $+31.79\%$ at $\alpha{=}0.25$).

\begin{figure}[h]
\centering
\includegraphics[width=\textwidth]{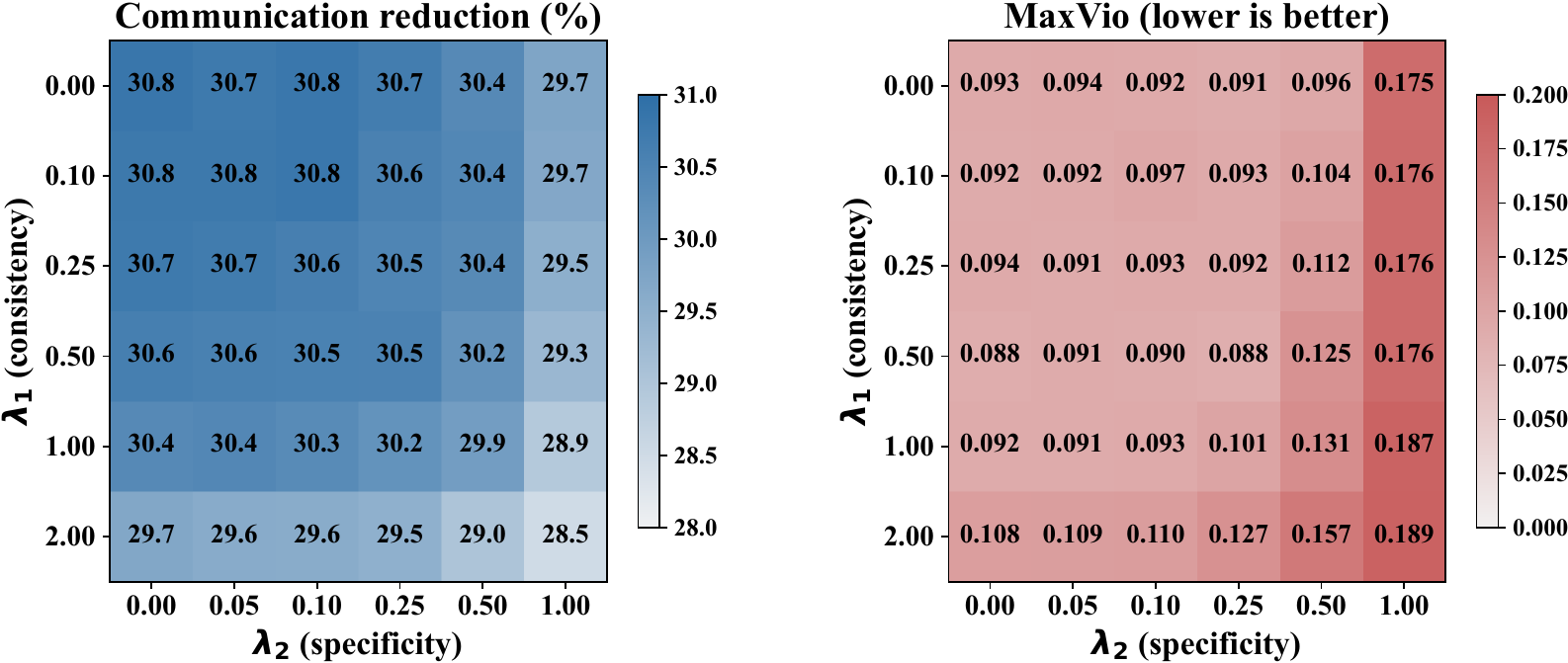}
\caption{Cross-model sensitivity to the GESR coefficients $(\lambda_1, \lambda_2)$. Rows index $\lambda_1$ (consistency reward), columns index $\lambda_2$ (specificity penalty). Left: communication reduction (higher is better). Right: MaxVio (lower is better). The canonical setting reported in the main paper is $(\lambda_1,\lambda_2)=(0,0)$.}
\label{fig:sweep-gesr-lambdas}
\end{figure}

\subsection{GESR Scoring Coefficients $(\lambda_1, \lambda_2)$}
\label{app:sweep_lambdas}

GESR scores generic experts by $r_{l,e} = \mathrm{Cent}_{l,e} + \lambda_1\,\mathrm{Cons}_{l,e} - \lambda_2\,\mathrm{Spec}_{l,e}$, where $\lambda_1$ rewards cross-family consistency and $\lambda_2$ penalizes family-specific deviation. We sweep $\lambda_1\in\{0, 0.1, 0.25, 0.5, 1.0, 2.0\}$ and $\lambda_2\in\{0, 0.05, 0.1, 0.25, 0.5, 1.0\}$, holding the replica budget fixed at $n_{\mathrm{rep}}{=}8$, $k_{\mathrm{sec}}{=}2$. The extra-memory cost is identical across all cells ($+25.6\%$ on average), so we omit it.

\paragraph{Obs.\,A2: Centrality is the dominant signal; both regularizers are mild and self-limiting.}
Across the entire grid (Figure~\ref{fig:sweep-gesr-lambdas}), Comm.\ Red.\ stays within a narrow band of $28.5\%$--$30.8\%$ and MaxVio within $0.088$--$0.189$. Increasing $\lambda_2$ aggressively (column $\lambda_2{=}1.0$) is the single most damaging move: it suppresses generic experts that happen to skew toward one family, which removes useful replicas and therefore both \emph{lowers} Comm.\ Red.\ ($-1$ to $-2$ pp) and \emph{raises} MaxVio (up to $0.189$). Increasing $\lambda_1$ aggressively (row $\lambda_1{=}2$) similarly degrades Comm.\ Red.\ by $\sim\!1$ pp by promoting experts whose cross-family consistency is high but whose centrality is moderate. Both regularizers are best kept small: any $(\lambda_1,\lambda_2)\in[0,0.5]\times[0,0.25]$ is within $0.4$ pp of the optimum, justifying the canonical choice $(0,0)$ that uses pure centrality.

\subsection{Load-Guard Slack $\theta$}
\label{app:sweep_theta}

The serving-time load guard restricts secondary-card admission to candidates whose recent load is at most $(1+\theta)\bar{L}$, where $\bar{L}$ is the recent per-GPU mean (Section~\ref{sec:gesr}). Setting $\theta\!\to\!\infty$ disables the guard (equivalent to \emph{Ours w/o load guard} in the ablation table); setting $\theta\!=\!0$ admits only below-average GPUs and corresponds to the strictest balance regime.


\begin{figure}[h]
    \centering
    \begin{subfigure}[b]{0.45\linewidth}
        \centering
        \includegraphics[width=\linewidth]{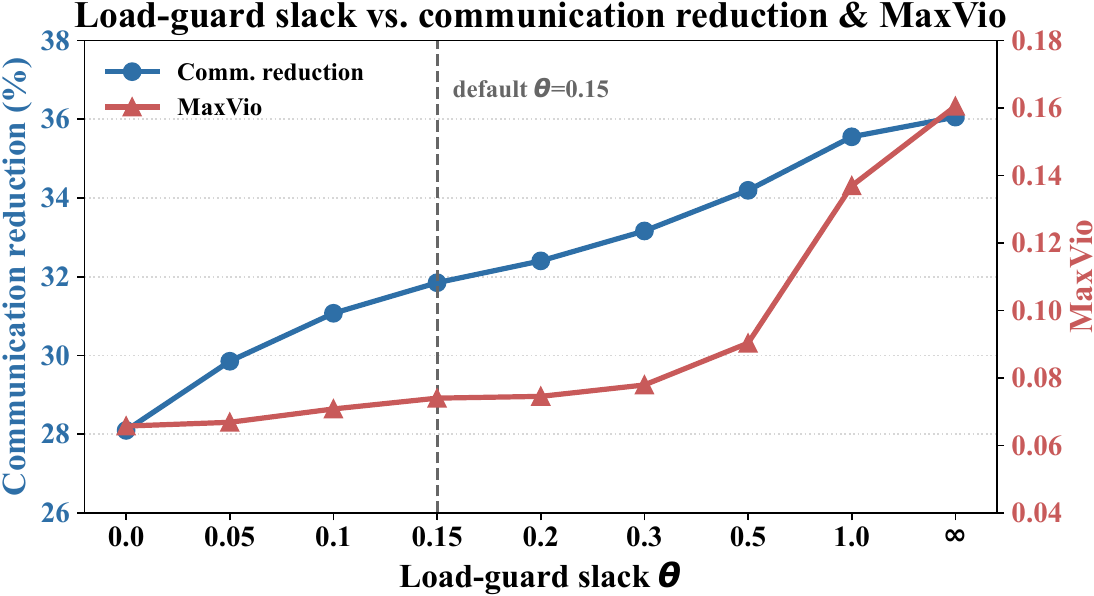}
        \label{fig:8}
    \end{subfigure}
    \hfill
    \begin{subfigure}[b]{0.45\linewidth}
        \centering
        \includegraphics[width=\linewidth]{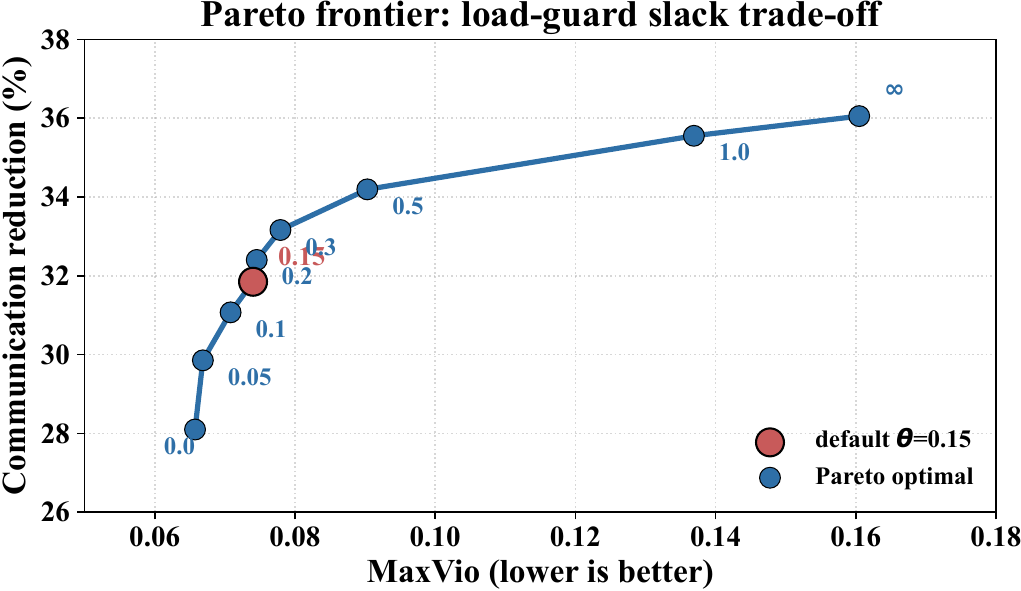}
        \label{fig:9}
    \end{subfigure}
    \caption{Cross-model sensitivity to the load-guard slack $\theta$. As $\theta$ grows, Comm.\ Red.\ rises monotonically while MaxVio degrades in the opposite direction; the canonical operating point $\theta{=}0.15$ captures most of the achievable communication gain at a still-tight MaxVio of $0.074$.}
    \label{fig:sweep-load-guard-theta}
\end{figure}

\paragraph{Obs.\,A3: The communication--balance trade-off is monotone and well-behaved in $\theta$.}
Comm.\ Red.\ rises monotonically from $+28.1\%$ at $\theta{=}0$ to $+36.1\%$ at $\theta{\to}\infty$, while MaxVio grows in the opposite direction from $0.066$ to $0.161$ (Figure~\ref{fig:sweep-load-guard-theta}). The frontier is therefore a single one-dimensional curve, and choosing $\theta$ amounts to choosing an operating point on it. We use $\theta=0.15$ in the main paper because it is the smallest slack that captures most of the available communication gain ($+31.85\%$, only $\sim\!4$ pp below the unguarded extreme) while keeping MaxVio at $0.074$, well below the $0.10$ threshold at which Jain starts to drop noticeably. Importantly, the rerouting rate stays close to $50\%$ across the entire sweep: the load guard reshapes \emph{which} secondary card is selected rather than throttling secondary use.

\section{Per-Token Communication Trace Analysis}
\label{app:comm_trace}

To confirm visually that the Comm.\ Red.\ gains in Section~\ref{sec:main_results} stem from \emph{task-aware} colocation rather than incidental graph structure, we compare per-token dispatch traces of \textsc{Ours} (TACG\,+\,GESR) against the strongest pooled-coactivation baseline (\textsc{Coactivation}) on DeepSeek-MoE-16B. Each token traverses $27$ MoE layers; under a fixed expert-to-GPU grouping, every layer dispatches its top-$k$ experts to one or more GPUs, which we record as: (i) \emph{cross-group dispatch count} per token, $\sum_{l}\max(0, |\mathrm{cards}_l|-1)$, equal to the \texttt{cross\_group\_comm\_count} of the main table; (ii) \emph{cross-family layers} per token, the number of layers whose dispatch set spans more than one family; and (iii) \emph{home-family dispatch-mass concentration}, the fraction of dispatch mass that stays within the four GPUs of the token's home family ($25\%$ uniform baseline). All three are measured per task family on the held-out test stream.

\begin{figure}[h]
\centering
\includegraphics[width=\textwidth]{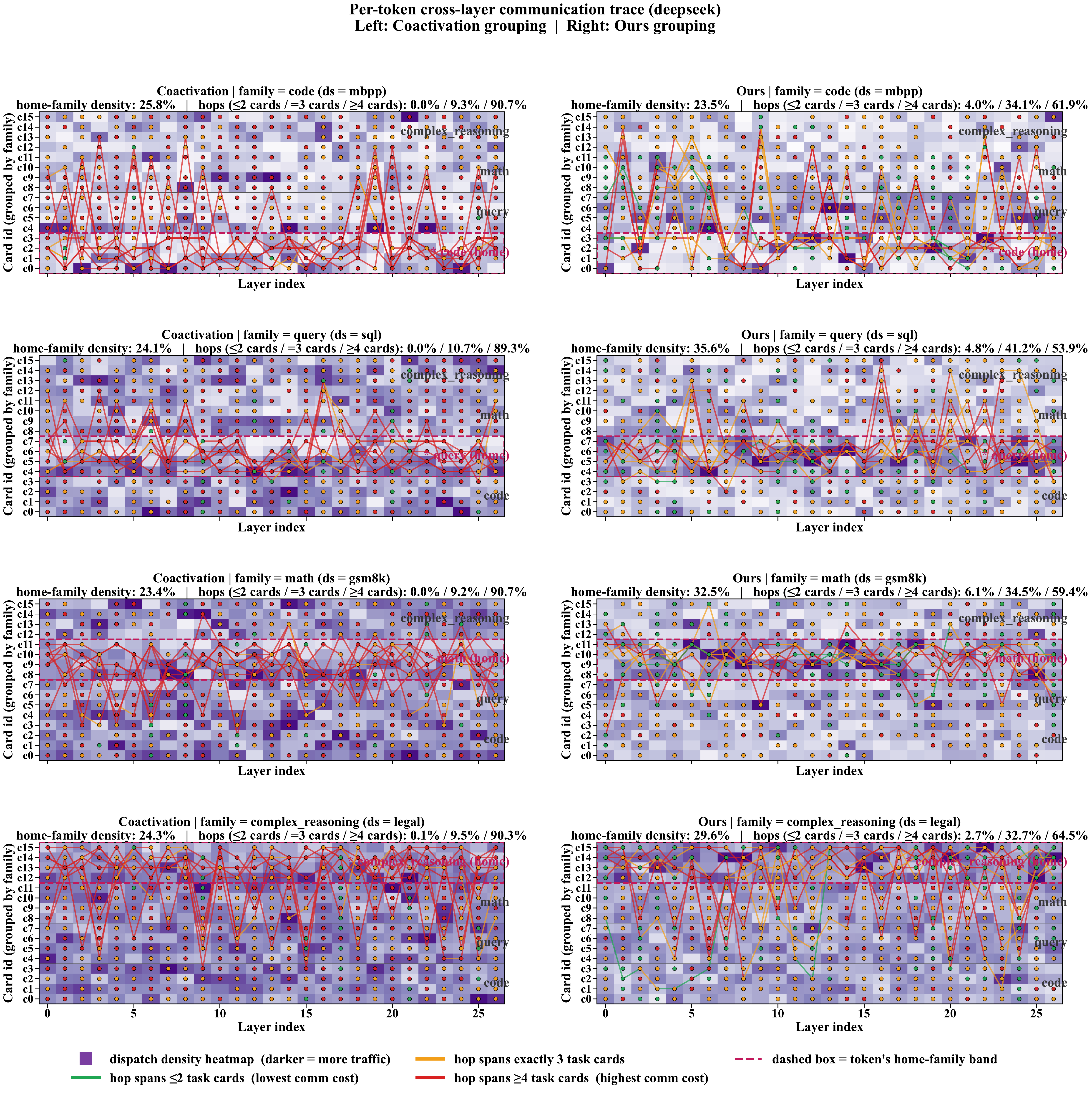}
\caption{Per-token communication trace of \textsc{Ours} vs.\ pooled \textsc{Coactivation} on DeepSeek-MoE-16B, evaluated on four task families (Code/MBPP, Query/SQL, Math/GSM8K, Complex/Legal). For each family the layered trace shows, for a representative token, the set of GPUs touched per MoE layer; \textsc{Ours} keeps a substantially larger fraction of layers concentrated within the home-family band.}
\label{fig:comm_trace}
\end{figure}

\begin{table}[h]
\centering
\small
\setlength{\tabcolsep}{4pt}
\caption{Per-token trace statistics of \textsc{Ours} vs.\ pooled \textsc{Coactivation} on DeepSeek-MoE-16B (lower is better for the first two columns; higher is better for the home-family concentration). $\Delta$ rows give the absolute or relative improvement of \textsc{Ours} over \textsc{Coactivation}.}
\label{tab:comm_trace}
\begin{tabular}{lrrrrrrrr}
\toprule
 & \multicolumn{2}{c}{\textbf{Cross-group hops/token}} & \multicolumn{2}{c}{\textbf{Cross-family layers/token}} & \multicolumn{3}{c}{\textbf{Home-family mass (\%)}} \\
\cmidrule(lr){2-3}\cmidrule(lr){4-5}\cmidrule(lr){6-8}
Family / Dataset & Coact & Ours & Coact & Ours & Coact & Ours & $\Delta$ (pp) \\
\midrule
Code / MBPP        & 101.10 & \textbf{84.58} ($-16.4\%$) & 26.99 & \textbf{26.42} & 26.24 & 24.64 & $-1.59$ \\
Query / SQL        & 102.01 & \textbf{82.56} ($-19.1\%$) & 26.99 & \textbf{26.21} & 24.29 & \textbf{34.48} & $+10.19$ \\
Math / GSM8K       & 102.18 & \textbf{83.80} ($-18.0\%$) & 26.97 & \textbf{26.25} & 23.61 & \textbf{31.94} & $+8.32$ \\
Complex / Legal    & 103.57 & \textbf{87.76} ($-15.3\%$) & 26.98 & \textbf{26.57} & 24.54 & \textbf{28.57} & $+4.02$ \\
\midrule
\textbf{Average}   & 102.22 & \textbf{84.68} ($-17.2\%$) & 26.98 & \textbf{26.36} & 24.67 & \textbf{29.91} & $+5.24$ \\
\bottomrule
\end{tabular}
\end{table}

\paragraph{Obs.\,A4: \textsc{Ours} cuts per-layer cross-group hops uniformly across all four families.}
\textsc{Ours} reduces the per-token cross-group hop count by $15$--$19\%$ on every family (Table~\ref{tab:comm_trace}), an improvement that closely tracks the headline Comm.\ Red.\ in Section~\ref{sec:main_results} and confirms that the system-level gains compose directly from per-layer dispatch reductions rather than from better load balance alone.


\section{Additional Experimental Setup}
\label{app:exp_setup}

\paragraph{Calibration corpora.}
For each task family in Section~\ref{sec:exp_setup} we draw calibration tokens from a small set of widely used open datasets: \texttt{mbpp}, \texttt{humaneval}, \texttt{apps-intro} for \texttt{code}; \texttt{sql}, \texttt{spider} for \texttt{query}; \texttt{gsm8k}, \texttt{numina-math} for \texttt{math}; and \texttt{bbh}, \texttt{legal} for \texttt{complex\_reasoning}. Calibration tokens feed only the per-family usage and coactivation statistics, and are never reused as an evaluation stream. The complete introduction of each dataset is in Appendix~\ref{app:task_intro}.

\paragraph{Per-GPU expert capacities.}
The four per-family GPU groups are kept symmetric: each MoE layer is deployed on $M{=}16$ GPUs partitioned into four groups of four GPUs, one per task family. Per-GPU capacities saturate the per-family expert budgets exactly: $\{4,4,4,4\}$ for DeepSeek-MoE-16B and Moonlight-16B-A3B (both $E{=}64$ routed experts per layer), and $\{4,4,4,3\}$ for Qwen1.5-MoE-A2.7B ($E{=}60$).

\paragraph{Stress-workload suite for the skew sweep.}
The skew study in Section~\ref{sec:rq2_taskaware} sweeps $13$ online workloads that all reuse the \emph{same} calibrated placement: one \texttt{balanced} reference; eight single-family-heavy streams (four families $\times$ two skew shares $\{0.6, 0.8\}$); two temporal phase-shift workloads that alternate the dominant family over time; and two bursty workloads that inject a $15\%$ burst of \texttt{math} or \texttt{complex\_reasoning} tokens on top of the balanced base. Only the online token stream changes across workloads; the deployed expert--GPU assignment is held fixed.

\section{Models}
\label{app:model_intro}

We evaluate three representative open-source sparse MoE backbones whose router structure and expert granularity span the design space we wish to stress (Table~\ref{tab:moe-backbones}).


\begin{table}[h]
\centering
\small
\caption{Sparse MoE backbones used in our experiments. ``MoE layers'' counts the layers whose FFN is replaced by an MoE block; the remaining decoder layers are dense.}
\label{tab:moe-backbones}
\begin{tabular}{lccccc}
\toprule
Model & Decoder layers & MoE layers & Routed experts & Shared experts & Top-$k$ \\
\midrule
DeepSeek-MoE-16B      & 28 & 27 & 64 & 2 & 6 \\
Qwen1.5-MoE-A2.7B     & 24 & 24 & 60 & 4 & 4 \\
Moonlight-16B-A3B     & 27 & 26 & 64 & 2 & 6 \\
\bottomrule
\end{tabular}
\end{table}

\textbf{DeepSeek-MoE-16B}~\citep{dai-etal-2024-deepseekmoe} adopts \emph{fine-grained expert segmentation}, in which each FFN is split into a larger number of smaller experts to enrich the per-token expert combination, and \emph{shared-expert isolation}, in which two always-on experts capture common knowledge so that the routed experts can specialize. The model is trained from scratch on $2$T Chinese--English tokens. Each MoE layer activates the two shared experts plus the top-$6$ of $64$ routed experts.

\textbf{Qwen1.5-MoE-A2.7B}~\citep{bai2023qwentechnicalreport} is a $14.3$B-parameter MoE upcycled from the dense Qwen-1.8B backbone rather than trained from scratch, reportedly using $\sim\!25\%$ of the training resources of Qwen1.5-7B at comparable quality. Each MoE layer routes each token to top-$4$ of $60$ routed experts on top of $4$ always-on shared experts.

\textbf{Moonlight-16B-A3B}~\citep{liu2025muonscalablellmtraining} follows a DeepSeek-V3-style sparse architecture and is trained on $5.7$T tokens with the Muon optimizer. The first decoder layer is dense; each of the remaining $26$ MoE layers activates two shared experts plus top-$6$ of $64$ routed experts.

\section{Tasks and Datasets}
\label{app:task_intro}

\textbf{MBPP}~\citep{austin2021programsynthesislargelanguage} contains crowd-sourced Python programming problems paired with reference solutions and unit tests. It probes natural-language-to-code generation and basic program synthesis ability. 

\textbf{HumanEval}~\citep{chen2021evaluatinglargelanguagemodels} is also a Python code generation benchmark. Each problem usually contains a function signature, a natural language docstring, and hidden unit tests. It evaluates whether the model can generate functionally correct Python functions. 

\textbf{APPS-Intro}~\citep{hendrycks2021measuringcodingchallengecompetence} is the introductory subset of the APPS programming benchmark. Compared with HumanEval, APPS problems are usually closer to full programming contest-style questions, where the model needs to understand the problem statement and generate a complete program that passes input-output tests. Together, these datasets represent the \textbf{Code} family.

\textbf{Spider}~\citep{yu2019spiderlargescalehumanlabeleddataset} is a cross-domain text-to-SQL benchmark whose multi-table schemas and nested queries probe structured-query generation. It evaluates whether a model can map a natural language question and a database schema to an executable SQL query. 

\textbf{SQL}~\citep{zhongSeq2SQL2017} refers to a general text-to-SQL dataset used for calibration in this work. It contains natural language questions, database contexts or schemas, and corresponding SQL queries. It is used to capture routing patterns related to structured query generation. Together, Spider and SQL represent the \textbf{Query} family.

\textbf{GSM8K}~\citep{cobbe2021trainingverifierssolvemath} comprises human-written grade-school math word problems whose solutions require multi-step arithmetic reasoning. It mainly tests whether a model can extract quantitative relations from natural language and solve them through step-by-step calculation. 

\textbf{Numina-Math}~\citep{LiNuminaMathTL} is a larger mathematical reasoning dataset that covers a wider range of math problems and solution styles. It is useful for collecting calibration traces that reflect more diverse mathematical reasoning patterns. Together, GSM8K and Numina-Math represent the \textbf{Math} family.

\textbf{Legal}~\citep{NEURIPS2023_89e44582} contains long, terminology-dense passages drawn from statutes, contracts, and case law, and probes long-context professional reasoning. 

\textbf{BBH}~\citep{suzgun-etal-2023-challenging}, or BIG-Bench Hard, is a collection of challenging reasoning tasks selected from BIG-Bench. It covers tasks that require multi-step reasoning, symbolic manipulation, logical inference, and complex language understanding. We use BBH together with Legal to represent the \textbf{Complex Reasoning} family, since both datasets require more complex reasoning paths than standard generation tasks. The four task families jointly cover code generation, structured generation, numeric reasoning, and long-context or complex professional reasoning, four regimes whose expert activation patterns are expected to differ substantially.

\section{Baseline Methods}
\label{app:baseline_methods}

\textbf{ExFlow}~\citep{MLSYS2023_5616d34c} accelerates pre-trained MoE inference by exploiting the implicit \emph{inter-layer expert affinity}: experts that frequently fire together across consecutive layers are colocated, and a context-coherent expert-parallel scheme keeps token context on its current GPU so that one All-to-All per layer is sufficient. ExFlow optimizes only placement and requires no fine-tuning.

\textbf{Semantic Parallelism}~\citep{li2026semantic} performs offline token-to-expert profiling, clusters experts that are activated by semantically similar tokens onto the same device, and then schedules online traffic accordingly: request-level scheduling under attention-DP and token-level reshuffling under attention-TP, so that tokens are moved in advance toward the devices that are more likely to host their target experts.

\textbf{MoETuner}~\citep{go2025moetuneroptimizedmixtureexpert} formulates expert placement as an integer linear program over a profiled \emph{cross-layer token routing graph}. The objective jointly minimizes per-GPU parameter cost, dispatch load imbalance, cross-GPU communication volume, and tail latency, exploiting the empirical observation that a token routed to a particular expert at one layer is likely routed to a small subset of experts at the next.

\textbf{C2R}~\citep{zhang-etal-2025-advancing} (\emph{Collaboration-Constrained Routing}) builds a per-layer expert-collaboration matrix from profiling, and at routing time first picks the top-$1$ expert and then restricts the remaining $K{-}1$ slots to the experts most often coactivated with it. Coactivated experts are colocated, so that one All-to-All hop suffices to deliver a token to its full expert subset on the local GPU.

\section{Metric Definitions}
\label{app:metric_intro}

\paragraph{System-level metrics.}
We use four system-level metrics that jointly characterize the communication--load-balance trade-off of an MoE deployment.

\textbf{Comm} is the per-token cross-GPU dispatch volume during MoE inference: every layer in which a token's selected experts span more than one GPU contributes one cross-GPU hop. Lower is better.

\textbf{Comm. Red.} (Communication Reduction) reports the relative drop in Comm against the vendor default placement baseline,
\begin{equation}
\mathrm{Comm.\ Red.} =
\frac{\mathrm{Comm}_{\mathrm{Baseline}} - \mathrm{Comm}_{\mathrm{Method}}}{\mathrm{Comm}_{\mathrm{Baseline}}}\times 100\%.
\label{eq:comm-red}
\end{equation}
Higher is better.

\textbf{Jain} is Jain's Fairness Index over the per-GPU dispatch loads $\{L_m\}_{m=1}^{M}$,
\begin{equation}
\mathrm{Jain} =
\frac{\left(\sum_{m=1}^{M} L_m\right)^2}{M\sum_{m=1}^{M} L_m^2} \in (0,1],
\label{eq:jain}
\end{equation}
which is a global measure of load uniformity (closer to $1$ is better).

\textbf{MaxVio} (Maximum Load Violation) measures the relative overload of the worst-case GPU,
\begin{equation}
\mathrm{MaxVio} = \frac{L_{\max} - \bar{L}}{\bar{L}},
\qquad
\bar{L}=\frac{1}{M}\sum_m L_m .
\label{eq:maxvio}
\end{equation}
Unlike Jain, MaxVio is dominated by the most overloaded device and therefore tracks the bottleneck risk of the system. Lower is better.

\textbf{$\Delta$Comm} reports the additive change in Comm. Red. of an ablation variant relative to the full method. A negative value means removing the module hurts; a positive value alone is \emph{not} a sign of improvement, since the variant may have traded load balance for raw communication reduction (e.g., \emph{w/o load guard} reaches $+35.87\%$ Comm. Red. but inflates MaxVio to $0.1422$, vs.\ the full method at $+31.39\%$ and $0.0736$).

\paragraph{Auxiliary motivation metrics.}
To substantiate the claim that expert coactivation is task-conditioned (Section~\ref{sec:intro}), we report three auxiliary diagnostics on calibration tokens. \textbf{Per-task expert activation} is the per-expert activation probability under each task family, used to visualize task-specific expert preferences. The \textbf{activation Frobenius distance} between two task families is the Frobenius norm $\|P_i - P_j\|_F$ of the difference between their expert-activation distributions; larger values indicate that the two families exercise different experts. The \textbf{top-$64$ pair Jaccard overlap} measures the Jaccard similarity between the sets of top-$64$ most frequently coactivated expert pairs of two task families; lower values indicate that the two families exercise different expert--expert collaboration structures. Together these diagnostics show that a globally pooled affinity graph hides task-specific coactivation patterns and motivates the task-aware grouping in Section~\ref{sec:primary_placement}.


\newpage
\section*{NeurIPS Paper Checklist}

The checklist is designed to encourage best practices for responsible machine learning research, addressing issues of reproducibility, transparency, research ethics, and societal impact. Do not remove the checklist: {\bf The papers not including the checklist will be desk rejected.} The checklist should follow the references and follow the (optional) supplemental material.  The checklist does NOT count towards the page
limit. 

Please read the checklist guidelines carefully for information on how to answer these questions. For each question in the checklist:
\begin{itemize}
    \item You should answer \answerYes{}, \answerNo{}, or \answerNA{}.
    \item \answerNA{} means either that the question is Not Applicable for that particular paper or the relevant information is Not Available.
    \item Please provide a short (1--2 sentence) justification right after your answer (even for \answerNA). 
\end{itemize}

{\bf The checklist answers are an integral part of your paper submission.} They are visible to the reviewers, area chairs, senior area chairs, and ethics reviewers. You will also be asked to include it (after eventual revisions) with the final version of your paper, and its final version will be published with the paper.

The reviewers of your paper will be asked to use the checklist as one of the factors in their evaluation. While \answerYes{} is generally preferable to \answerNo{}, it is perfectly acceptable to answer \answerNo{} provided a proper justification is given (e.g., error bars are not reported because it would be too computationally expensive'' or ``we were unable to find the license for the dataset we used''). In general, answering \answerNo{} or \answerNA{} is not grounds for rejection. While the questions are phrased in a binary way, we acknowledge that the true answer is often more nuanced, so please just use your best judgment and write a justification to elaborate. All supporting evidence can appear either in the main paper or the supplemental material, provided in appendix. If you answer \answerYes{} to a question, in the justification please point to the section(s) where related material for the question can be found.

IMPORTANT, please:
\begin{itemize}
    \item {\bf Delete this instruction block, but keep the section heading ``NeurIPS Paper Checklist"},
    \item  {\bf Keep the checklist subsection headings, questions/answers and guidelines below.}
    \item {\bf Do not modify the questions and only use the provided macros for your answers}.
\end{itemize}


\begin{enumerate}

\item {\bf Claims}
    \item[] Question: Do the main claims made in the abstract and introduction accurately reflect the paper's contributions and scope?
    \item[] Answer: \answerYes{}
    \item[] Justification: We include the paper’s main claims in the abstract and introduction.
    \item[] Guidelines:
    \begin{itemize}
        \item The answer \answerNA{} means that the abstract and introduction do not include the claims made in the paper.
        \item The abstract and/or introduction should clearly state the claims made, including the contributions made in the paper and important assumptions and limitations. A \answerNo{} or \answerNA{} answer to this question will not be perceived well by the reviewers. 
        \item The claims made should match theoretical and experimental results, and reflect how much the results can be expected to generalize to other settings. 
        \item It is fine to include aspirational goals as motivation as long as it is clear that these goals are not attained by the paper. 
    \end{itemize}

\item {\bf Limitations}
    \item[] Question: Does the paper discuss the limitations of the work performed by the authors?
    \item[] Answer: \answerYes{} 
    \item[] Justification: We discuss the limitations of our work in Section~\ref{sec:limit}.
    \item[] Guidelines:
    \begin{itemize}
        \item The answer \answerNA{} means that the paper has no limitation while the answer \answerNo{} means that the paper has limitations, but those are not discussed in the paper. 
        \item The authors are encouraged to create a separate ``Limitations'' section in their paper.
        \item The paper should point out any strong assumptions and how robust the results are to violations of these assumptions (e.g., independence assumptions, noiseless settings, model well-specification, asymptotic approximations only holding locally). The authors should reflect on how these assumptions might be violated in practice and what the implications would be.
        \item The authors should reflect on the scope of the claims made, e.g., if the approach was only tested on a few datasets or with a few runs. In general, empirical results often depend on implicit assumptions, which should be articulated.
        \item The authors should reflect on the factors that influence the performance of the approach. For example, a facial recognition algorithm may perform poorly when image resolution is low or images are taken in low lighting. Or a speech-to-text system might not be used reliably to provide closed captions for online lectures because it fails to handle technical jargon.
        \item The authors should discuss the computational efficiency of the proposed algorithms and how they scale with dataset size.
        \item If applicable, the authors should discuss possible limitations of their approach to address problems of privacy and fairness.
        \item While the authors might fear that complete honesty about limitations might be used by reviewers as grounds for rejection, a worse outcome might be that reviewers discover limitations that aren't acknowledged in the paper. The authors should use their best judgment and recognize that individual actions in favor of transparency play an important role in developing norms that preserve the integrity of the community. Reviewers will be specifically instructed to not penalize honesty concerning limitations.
    \end{itemize}

\item {\bf Theory assumptions and proofs}
    \item[] Question: For each theoretical result, does the paper provide the full set of assumptions and a complete (and correct) proof?
    \item[] Answer: \answerNA{}
    \item[] Justification: This paper does not include theoretical results.
    \item[] Guidelines:
    \begin{itemize}
        \item The answer \answerNA{} means that the paper does not include theoretical results. 
        \item All the theorems, formulas, and proofs in the paper should be numbered and cross-referenced.
        \item All assumptions should be clearly stated or referenced in the statement of any theorems.
        \item The proofs can either appear in the main paper or the supplemental material, but if they appear in the supplemental material, the authors are encouraged to provide a short proof sketch to provide intuition. 
        \item Inversely, any informal proof provided in the core of the paper should be complemented by formal proofs provided in appendix or supplemental material.
        \item Theorems and Lemmas that the proof relies upon should be properly referenced. 
    \end{itemize}

    \item {\bf Experimental result reproducibility}
    \item[] Question: Does the paper fully disclose all the information needed to reproduce the main experimental results of the paper to the extent that it affects the main claims and/or conclusions of the paper (regardless of whether the code and data are provided or not)?
    \item[] Answer: \answerYes{} 
    \item[] Justification: We provide the code of our paper in the supplementary material and anonymous links to support the reproducibility of the main experimental results.
    \item[] Guidelines:
    \begin{itemize}
        \item The answer \answerNA{} means that the paper does not include experiments.
        \item If the paper includes experiments, a \answerNo{} answer to this question will not be perceived well by the reviewers: Making the paper reproducible is important, regardless of whether the code and data are provided or not.
        \item If the contribution is a dataset and\slash or model, the authors should describe the steps taken to make their results reproducible or verifiable. 
        \item Depending on the contribution, reproducibility can be accomplished in various ways. For example, if the contribution is a novel architecture, describing the architecture fully might suffice, or if the contribution is a specific model and empirical evaluation, it may be necessary to either make it possible for others to replicate the model with the same dataset, or provide access to the model. In general. releasing code and data is often one good way to accomplish this, but reproducibility can also be provided via detailed instructions for how to replicate the results, access to a hosted model (e.g., in the case of a large language model), releasing of a model checkpoint, or other means that are appropriate to the research performed.
        \item While NeurIPS does not require releasing code, the conference does require all submissions to provide some reasonable avenue for reproducibility, which may depend on the nature of the contribution. For example
        \begin{enumerate}
            \item If the contribution is primarily a new algorithm, the paper should make it clear how to reproduce that algorithm.
            \item If the contribution is primarily a new model architecture, the paper should describe the architecture clearly and fully.
            \item If the contribution is a new model (e.g., a large language model), then there should either be a way to access this model for reproducing the results or a way to reproduce the model (e.g., with an open-source dataset or instructions for how to construct the dataset).
            \item We recognize that reproducibility may be tricky in some cases, in which case authors are welcome to describe the particular way they provide for reproducibility. In the case of closed-source models, it may be that access to the model is limited in some way (e.g., to registered users), but it should be possible for other researchers to have some path to reproducing or verifying the results.
        \end{enumerate}
    \end{itemize}

\item {\bf Open access to data and code}
    \item[] Question: Does the paper provide open access to the data and code, with sufficient instructions to faithfully reproduce the main experimental results, as described in supplemental material?
    \item[] Answer: \answerYes{} 
    \item[] Justification: We provide the code in the supplementary material and anonymous links, together with instructions for reproducing the main experimental results.
    \item[] Guidelines:
    \begin{itemize}
        \item The answer \answerNA{} means that paper does not include experiments requiring code.
        \item Please see the NeurIPS code and data submission guidelines (\url{https://neurips.cc/public/guides/CodeSubmissionPolicy}) for more details.
        \item While we encourage the release of code and data, we understand that this might not be possible, so \answerNo{} is an acceptable answer. Papers cannot be rejected simply for not including code, unless this is central to the contribution (e.g., for a new open-source benchmark).
        \item The instructions should contain the exact command and environment needed to run to reproduce the results. See the NeurIPS code and data submission guidelines (\url{https://neurips.cc/public/guides/CodeSubmissionPolicy}) for more details.
        \item The authors should provide instructions on data access and preparation, including how to access the raw data, preprocessed data, intermediate data, and generated data, etc.
        \item The authors should provide scripts to reproduce all experimental results for the new proposed method and baselines. If only a subset of experiments are reproducible, they should state which ones are omitted from the script and why.
        \item At submission time, to preserve anonymity, the authors should release anonymized versions (if applicable).
        \item Providing as much information as possible in supplemental material (appended to the paper) is recommended, but including URLs to data and code is permitted.
    \end{itemize}

\item {\bf Experimental setting/details}
    \item[] Question: Does the paper specify all the training and test details (e.g., data splits, hyperparameters, how they were chosen, type of optimizer) necessary to understand the results?
    \item[] Answer: \answerYes{}  
    \item[] Justification: We specify the experimental settings/details in Section~\ref{sec:exp_setup} and Appendix~\ref{app:exp_setup}.
    \item[] Guidelines:
    \begin{itemize}
        \item The answer \answerNA{} means that the paper does not include experiments.
        \item The experimental setting should be presented in the core of the paper to a level of detail that is necessary to appreciate the results and make sense of them.
        \item The full details can be provided either with the code, in appendix, or as supplemental material.
    \end{itemize}

\item {\bf Experiment statistical significance}
    \item[] Question: Does the paper report error bars suitably and correctly defined or other appropriate information about the statistical significance of the experiments?
    \item[] Answer: \answerYes{} 
    \item[] Justification:We report the statistical significance analysis in Section~\ref{sec:experiments}.
    \item[] Guidelines:
    \begin{itemize}
        \item The answer \answerNA{} means that the paper does not include experiments.
        \item The authors should answer \answerYes{} if the results are accompanied by error bars, confidence intervals, or statistical significance tests, at least for the experiments that support the main claims of the paper.
        \item The factors of variability that the error bars are capturing should be clearly stated (for example, train/test split, initialization, random drawing of some parameter, or overall run with given experimental conditions).
        \item The method for calculating the error bars should be explained (closed form formula, call to a library function, bootstrap, etc.)
        \item The assumptions made should be given (e.g., Normally distributed errors).
        \item It should be clear whether the error bar is the standard deviation or the standard error of the mean.
        \item It is OK to report 1-sigma error bars, but one should state it. The authors should preferably report a 2-sigma error bar than state that they have a 96\% CI, if the hypothesis of Normality of errors is not verified.
        \item For asymmetric distributions, the authors should be careful not to show in tables or figures symmetric error bars that would yield results that are out of range (e.g., negative error rates).
        \item If error bars are reported in tables or plots, the authors should explain in the text how they were calculated and reference the corresponding figures or tables in the text.
    \end{itemize}

\item {\bf Experiments compute resources}
    \item[] Question: For each experiment, does the paper provide sufficient information on the computer resources (type of compute workers, memory, time of execution) needed to reproduce the experiments?
    \item[] Answer:  \answerYes{}  
    \item[] Justification:  We report the average running time per sample in Section~\ref{sec:exp_setup}.
    \item[] Guidelines:
    \begin{itemize}
        \item The answer \answerNA{} means that the paper does not include experiments.
        \item The paper should indicate the type of compute workers CPU or GPU, internal cluster, or cloud provider, including relevant memory and storage.
        \item The paper should provide the amount of compute required for each of the individual experimental runs as well as estimate the total compute. 
        \item The paper should disclose whether the full research project required more compute than the experiments reported in the paper (e.g., preliminary or failed experiments that didn't make it into the paper). 
    \end{itemize}
    
\item {\bf Code of ethics}
    \item[] Question: Does the research conducted in the paper conform, in every respect, with the NeurIPS Code of Ethics \url{https://neurips.cc/public/EthicsGuidelines}?
    \item[] Answer: \answerYes{} 
    \item[] Justification:  The research conducted in this paper conforms to the NeurIPS Code of Ethics.
    \item[] Guidelines:
    \begin{itemize}
        \item The answer \answerNA{} means that the authors have not reviewed the NeurIPS Code of Ethics.
        \item If the authors answer \answerNo, they should explain the special circumstances that require a deviation from the Code of Ethics.
        \item The authors should make sure to preserve anonymity (e.g., if there is a special consideration due to laws or regulations in their jurisdiction).
    \end{itemize}

\item {\bf Broader impacts}
    \item[] Question: Does the paper discuss both potential positive societal impacts and negative societal impacts of the work performed?
    \item[] Answer: \answerYes{}  
    \item[] Justification: We discuss the potential societal impacts of improving automated vulnerability detection in Section~\ref{sec:limit}.
    \item[] Guidelines:
    \begin{itemize}
        \item The answer \answerNA{} means that there is no societal impact of the work performed.
        \item If the authors answer \answerNA{} or \answerNo, they should explain why their work has no societal impact or why the paper does not address societal impact.
        \item Examples of negative societal impacts include potential malicious or unintended uses (e.g., disinformation, generating fake profiles, surveillance), fairness considerations (e.g., deployment of technologies that could make decisions that unfairly impact specific groups), privacy considerations, and security considerations.
        \item The conference expects that many papers will be foundational research and not tied to particular applications, let alone deployments. However, if there is a direct path to any negative applications, the authors should point it out. For example, it is legitimate to point out that an improvement in the quality of generative models could be used to generate Deepfakes for disinformation. On the other hand, it is not needed to point out that a generic algorithm for optimizing neural networks could enable people to train models that generate Deepfakes faster.
        \item The authors should consider possible harms that could arise when the technology is being used as intended and functioning correctly, harms that could arise when the technology is being used as intended but gives incorrect results, and harms following from (intentional or unintentional) misuse of the technology.
        \item If there are negative societal impacts, the authors could also discuss possible mitigation strategies (e.g., gated release of models, providing defenses in addition to attacks, mechanisms for monitoring misuse, mechanisms to monitor how a system learns from feedback over time, improving the efficiency and accessibility of ML).
    \end{itemize}
    
\item {\bf Safeguards}
    \item[] Question: Does the paper describe safeguards that have been put in place for responsible release of data or models that have a high risk for misuse (e.g., pre-trained language models, image generators, or scraped datasets)?
    \item[] Answer: \answerNA{} 
    \item[] Justification: This paper does not release high-risk pretrained models, image generators, or scraped datasets that require additional safeguards for responsible release.
    \item[] Guidelines:
    \begin{itemize}
        \item The answer \answerNA{} means that the paper poses no such risks.
        \item Released models that have a high risk for misuse or dual-use should be released with necessary safeguards to allow for controlled use of the model, for example by requiring that users adhere to usage guidelines or restrictions to access the model or implementing safety filters. 
        \item Datasets that have been scraped from the Internet could pose safety risks. The authors should describe how they avoided releasing unsafe images.
        \item We recognize that providing effective safeguards is challenging, and many papers do not require this, but we encourage authors to take this into account and make a best faith effort.
    \end{itemize}

\item {\bf Licenses for existing assets}
    \item[] Question: Are the creators or original owners of assets (e.g., code, data, models), used in the paper, properly credited and are the license and terms of use explicitly mentioned and properly respected?
    \item[] Answer:\answerYes{}
    \item[] Justification:  We provide asset licenses and usage terms in the supplementary README.
    \item[] Guidelines:
    \begin{itemize}
        \item The answer \answerNA{} means that the paper does not use existing assets.
        \item The authors should cite the original paper that produced the code package or dataset.
        \item The authors should state which version of the asset is used and, if possible, include a URL.
        \item The name of the license (e.g., CC-BY 4.0) should be included for each asset.
        \item For scraped data from a particular source (e.g., website), the copyright and terms of service of that source should be provided.
        \item If assets are released, the license, copyright information, and terms of use in the package should be provided. For popular datasets, \url{paperswithcode.com/datasets} has curated licenses for some datasets. Their licensing guide can help determine the license of a dataset.
        \item For existing datasets that are re-packaged, both the original license and the license of the derived asset (if it has changed) should be provided.
        \item If this information is not available online, the authors are encouraged to reach out to the asset's creators.
    \end{itemize}

\item {\bf New assets}
    \item[] Question: Are new assets introduced in the paper well documented and is the documentation provided alongside the assets?
    \item[] Answer: \answerYes{} 
    \item[] Justification: We provide the code with documentation in the anonymous links.
    \item[] Guidelines:
    \begin{itemize}
        \item The answer \answerNA{} means that the paper does not release new assets.
        \item Researchers should communicate the details of the dataset\slash code\slash model as part of their submissions via structured templates. This includes details about training, license, limitations, etc. 
        \item The paper should discuss whether and how consent was obtained from people whose asset is used.
        \item At submission time, remember to anonymize your assets (if applicable). You can either create an anonymized URL or include an anonymized zip file.
    \end{itemize}

\item {\bf Crowdsourcing and research with human subjects}
    \item[] Question: For crowdsourcing experiments and research with human subjects, does the paper include the full text of instructions given to participants and screenshots, if applicable, as well as details about compensation (if any)? 
    \item[] Answer: \answerNA{} 
    \item[] Justification: This paper does not involve crowdsourcing or research with human subjects.
    \item[] Guidelines:
    \begin{itemize}
        \item The answer \answerNA{} means that the paper does not involve crowdsourcing nor research with human subjects.
        \item Including this information in the supplemental material is fine, but if the main contribution of the paper involves human subjects, then as much detail as possible should be included in the main paper. 
        \item According to the NeurIPS Code of Ethics, workers involved in data collection, curation, or other labor should be paid at least the minimum wage in the country of the data collector. 
    \end{itemize}

\item {\bf Institutional review board (IRB) approvals or equivalent for research with human subjects}
    \item[] Question: Does the paper describe potential risks incurred by study participants, whether such risks were disclosed to the subjects, and whether Institutional Review Board (IRB) approvals (or an equivalent approval/review based on the requirements of your country or institution) were obtained?
    \item[] Answer: \answerNA{}
    \item[] Justification:This paper does not involve crowdsourcing or research with human subjects.
    \item[] Guidelines:
    \begin{itemize}
        \item The answer \answerNA{} means that the paper does not involve crowdsourcing nor research with human subjects.
        \item Depending on the country in which research is conducted, IRB approval (or equivalent) may be required for any human subjects research. If you obtained IRB approval, you should clearly state this in the paper. 
        \item We recognize that the procedures for this may vary significantly between institutions and locations, and we expect authors to adhere to the NeurIPS Code of Ethics and the guidelines for their institution. 
        \item For initial submissions, do not include any information that would break anonymity (if applicable), such as the institution conducting the review.
    \end{itemize}

\item {\bf Declaration of LLM usage}
    \item[] Question: Does the paper describe the usage of LLMs if it is an important, original, or non-standard component of the core methods in this research? Note that if the LLM is used only for writing, editing, or formatting purposes and does \emph{not} impact the core methodology, scientific rigor, or originality of the research, declaration is not required.
    \item[] Answer: \answerYes{}  
    \item[] Justification: We describe the LLM usage in the method and experiments.
    \item[] Guidelines:
    \begin{itemize}
        \item The answer \answerNA{} means that the core method development in this research does not involve LLMs as any important, original, or non-standard components.
        \item Please refer to our LLM policy in the NeurIPS handbook for what should or should not be described.
    \end{itemize}

\end{enumerate}

\end{document}